\documentclass[10pt,journal,compsoc]{IEEEtran}
\IEEEoverridecommandlockouts

\ifCLASSINFOpdf
\else
\fi

\usepackage{cite}
\usepackage{amsmath,amssymb,amsfonts}
\usepackage[noend,linesnumbered,algoruled,boxed,lined]{algorithm2e}

\usepackage{graphicx}
\usepackage{textcomp}
\usepackage{xcolor}
\usepackage{multirow}
\usepackage{bm}
\usepackage{booktabs}
\usepackage{tabularx}
\usepackage[caption=false]{subfig}
\DeclareSubrefFormat{parens}{#1(#2)}
\usepackage{soul}

\newtheorem{definition}{\noindent \textbf{Definition}}

\allowdisplaybreaks[4]

\begin{document}

\title{Aerial Secure Collaborative Communications under Eavesdropper Collusion in Low-altitude Economy: A Generative Swarm Intelligent Approach}

	

\author{
Jiahui Li, \IEEEmembership{Member,~IEEE,}
Geng~Sun, \IEEEmembership{Senior Member,~IEEE,}
Qingqing~Wu, \IEEEmembership{Senior Member,~IEEE,}
Shuang~Liang,
Jiacheng~Wang,
Dusit Niyato, \IEEEmembership{Fellow,~IEEE,}
and Dong In Kim,~\IEEEmembership{Life Fellow,~IEEE}


  \thanks{This work is supported in part by the National Natural Science Foundation of China (62172186, 62272194, 62471200), in part by the Science and Technology Development Plan Project of Jilin Province (20230201087GX), in part by the Postdoctoral Fellowship Program of China Postdoctoral Science Foundation (GZC20240592), in part by China Postdoctoral Science Foundation General Fund (2024M761123), and in part by the Scientific Research Project of Jilin Provincial Department of Education (JJKH20250117KJ). \textit{(Corresponding author: Geng Sun.)}
  
  \par Jiahui Li is with the College of Computer Science and Technology, Jilin University, Changchun 130012, China (E-mail: lijiahui@jlu.edu.cn). 
  
  \par Geng Sun is with the College of Computer Science and Technology, Jilin University, Changchun 130012, China, and also with the Key Laboratory of Symbolic Computation and Knowledge Engineering of Ministry of Education, Jilin University, Changchun 130012, China. He is also with the College of Computing and Data Science, Nanyang Technological University, Singapore 639798 (E-mail: sungeng@jlu.edu.cn). 
  
  \par Qingqing Wu is with the Department of Electronic Engineering, Shanghai Jiao Tong University, Shanghai 200240, China  (E-mail: qingqingwu@sjtu.edu.cn). 

  \par Shuang Liang is with the School of Information Science and Technology, Northeast Normal University, Changchun, 130117, China, and also with Key Laboratory of Symbolic Computation and Knowledge Engineering of Ministry of Education, Jilin University, Changchun 130012, China (e-mail: liangshuang@nenu.edu.cn).
  
  \par Jiacheng Wang and Dusit Niyato are with the College of Computing and Data Science, Nanyang Technological University, Singapore 639798 (E-mails: jiacheng.wang@ntu.edu.sg; dniyato@ntu.edu.sg). 

   \par Dong In Kim is with the Department of Electrical and Computer Engineering, Sungkyunkwan University, Suwon 16419, South Korea. (E-mail:dongin@skku.edu).

  \par Part of this paper appeared in IEEE INFOCOM 2024~\cite{Li2024b}.
  }

}
	


\IEEEtitleabstractindextext{%
\begin{abstract}
The rapid development of the low-altitude economy (LAE) has significantly increased the utilization of autonomous aerial vehicles (AAVs) in various applications, necessitating efficient and secure communication methods among AAV swarms. In this work, we aim to introduce distributed collaborative beamforming (DCB) into AAV swarms and handle the eavesdropper collusion by controlling the corresponding signal distributions. Specifically, we consider a two-way DCB-enabled aerial communication between two AAV swarms and construct these swarms as two AAV virtual antenna arrays. Then, we minimize the two-way known secrecy capacity and maximum sidelobe level to avoid information leakage from the known and unknown eavesdroppers, respectively. Simultaneously, we also minimize the energy consumption of AAVs when constructing virtual antenna arrays. Due to the conflicting relationships between secure performance and energy efficiency, we consider these objectives by formulating a multi-objective optimization problem, which is NP-hard and with a large number of decision variables. Accordingly, we design a novel generative swarm intelligence (GenSI) framework to solve the problem with less overhead, which contains a conditional variational autoencoder (CVAE)-based generative method and a proposed powerful swarm intelligence algorithm. In this framework, CVAE can collect expert solutions obtained by the swarm intelligence algorithm in other environment states to explore characteristics and patterns, thereby directly generating high-quality initial solutions in new environment factors for the swarm intelligence algorithm to search solution space efficiently. Simulation results show that the proposed swarm intelligence algorithm outperforms other state-of-the-art baseline algorithms, and the GenSI can achieve similar optimization results by using far fewer iterations than the ordinary swarm intelligence algorithm. Experimental tests demonstrate that introducing the CVAE mechanism achieves a 58.7\% reduction in execution time, which enables the deployment of GenSI even on AAV platforms with limited computing power.
\end{abstract}

\begin{IEEEkeywords}
Distributed collaborative beamforming, eavesdropper collusion, multi-objective optimization, AAV secure communications, generative artificial intelligence.
\end{IEEEkeywords}
}

\maketitle
%

\IEEEdisplaynontitleabstractindextext
\IEEEpeerreviewmaketitle


%
%
\section{Introduction} 
\label{sec:introduction}

\par The low-altitude economy (LAE) represents an emerging frontier in modern economic development, which usually encompasses commercial activities in the airspace between ground level and 1,000 meters~\cite{yangEmbodiedAIempoweredLow2024}. Driven by technological advancements and regulatory reforms that have opened up previously underutilized airspace for commercial exploitation, this new economic sphere has witnessed exponential growth in recent years. Leading companies like DJI and EHang have demonstrated the diverse applications of this airspace~\cite{DAI2024108791}. Likewise, traditional logistics giants such as Amazon (Prime Air) have also entered this space by developing drone delivery networks to revolutionize last-mile delivery services~\cite{Jones2023}. With applications spanning aerial photography, agricultural monitoring, infrastructure inspection, and urban air mobility, the LAE is creating unprecedented opportunities across multiple sectors, with a projected market value expected to reach billions of dollars globally.

\par The rapid development of the LAE has significantly increased the utilization of autonomous aerial vehicles (AAVs), also referred to as drones, unmanned aerial vehicles, and electric vertical take-off and landing aircraft (eVTOLs), in various applications. Specifically, AAVs can be integrated with wireless communication functions to deliver various communication services~\cite{Dai2022}. Thanks to their advantages including high LoS probability, operational flexibility, and economic efficiency, AAVs play a vital role in next-generation communication systems and networks~\cite{Zhang2023}. For example, during emergency situations, AAVs can be rapidly deployed to serve as aerial relays supporting temporary ground networks, helping coordinate and enhance rescue operations~\cite{Prasad2023}. Moreover, AAVs possess the capability to collect data from distant or inaccessible monitoring locations and transmit them to data fusion centers, enabling flexible data acquisition~\cite{Ning2024}. In addition, AAVs can function as aerial base stations delivering coverage to ground users, thus providing demand-driven and economical network support~\cite{Abubakar2023}.


\par Among various applications propelled by the LAE, AAV-enabled air-to-air (A2A) communication is a promising platform in terrestrial network inadequate scenarios, \textit{e.g.,} urban package delivery, agricultural monitoring, and infrastructure inspection~\cite{Pandey2024}. Given that the reliable and secure operation of these services directly impacts the economic value of LAE, ensuring communication security becomes particularly crucial. Since the AAV transmitter and receiver are both at high altitudes, the multi-path effect caused by terrain is slight. Thus, AAV-enabled A2A communications are often with increased stability and throughput~\cite{Shi2022}. Nevertheless, the high LoS probability makes A2A communications susceptible to eavesdropping attacks, especially in scenarios involving colluding eavesdroppers~\cite{Cao2023}. Under such circumstances, multiple attackers can coordinate their actions to conceal their presence to avoid complete detection, making it difficult for aerial communication systems to defend against these threats. Traditional upper-layer encryption techniques are established methods for securing wireless communication links and ensuring confidentiality against eavesdroppers. However, implementing such computation-intensive encryption algorithms poses significant challenges for AAVs with limited hardware capabilities.

\par Physical layer security (PLS) emerges as a promising security solution that leverages physical characteristics of wireless channels, including channel fading randomness, to enable secure transmissions~\cite{Feng2023}. This approach achieves communication security without depending on sophisticated encryption algorithms or authentication protocols, making it well-suited for LAE scenarios, particularly in long-range communication scenarios. Given the high mobility characteristics of AAVs, many studies investigate various PLS-based methods to counter eavesdropping threats. However, current research primarily focuses on AAV trajectory design (\textit{e.g.},\cite{Maeng2022,Yapici2021,Yin2022}) and power allocation strategies (\textit{e.g.},\cite{Zhang2019,Na2022}), which introduce two significant challenges. \textit{First}, frequent trajectory adjustments increase the AAV energy consumption, consequently shortening the available service duration and degrading LAE operational efficiency, especially in scenarios requiring sustained communication. Similarly, power allocation methods restrict the transmit power of the AAVs, leading to extended transmission periods and elevated hovering energy requirements, which can be problematic in long-distance operations. \textit{Second}, these approaches struggle to effectively address eavesdropper collusion scenarios, since mishandling any single eavesdropper could compromise the entire security framework, particularly in environments with multiple potential threats. As such, it is desirable to investigate a novel PLS to avoid eavesdropper collusion in long-range A2A communications.

\par In this work, we propose to use distributed collaborative beamforming (DCB)~\cite{Sun2021a} to establish an effective security mechanism against eavesdropper collusion in LAE aerial communications. Consider a typical long-range two-way aerial communication between two AAV swarms, we construct each AAV swarm as an AAV virtual antenna array (AVAA). When the system suffers severe secure threats of eavesdropper collusion, we can carefully design the signal distributions (\textit{i.e.}, beam pattern) of AVAAs to suppress the signal strengths toward each detected eavesdropper simultaneously, which is crucial in long-range communication scenarios. Meanwhile, the excess signals from AVAA can be regulated to defend against potential undetected eavesdroppers. Through such precise signal manipulation, transmissions toward all eavesdroppers are minimized, thus successfully addressing the challenges posed by eavesdropper collusion.

\par Nevertheless, it is not straightforward to obtain qualified beam patterns for AVAAs. On the one hand, the beam patterns of AVAAs are affected by the positions and transmit powers of AAVs in a non-linear manner~\cite{Jayaprakasam2017}. Considering the environmental factors (such as the initial positions of the AAV swarms and eavesdroppers) changes, the existing methods based on conventional methods, such as swarm intelligence and non-convex optimization~\cite{Li2023,Wu2023}, need online computing to determine these decision variables, resulting in computational latency and runtime computational overhead, which can hinder real-time applications. On the other hand, when AAVs fine-tune their positions, the energy cost of AAVs will no doubt increase, particularly in scenarios where energy efficiency is critical. Thus, we should capture and balance the trade-offs between secure performance and energy efficiencies of AAVs in LAE. Accordingly, we aim to propose a computation-efficient multi-objective optimization method in DCB-enabled A2A communications under eavesdropper collusion. Our main contributions are listed as follows.

\begin{itemize}

  \item \textit{Novel Paradigm for Solving Multi-eavesdropper Collusion:} We consider a typical DCB-enabled long-range aerial two-way communication scenario of two AAV swarms under eavesdropper collusion for supporting diverse LAE applications, which security heavily relies on reliable A2A communications. Specifically, eavesdroppers collude based on signal detection, which leads to the worst wiretap case. In this case, we introduce DCB into each AAV swarm and use the signal processing method to handle eavesdroppers, aiming to ensure the secure and efficient operation of LAE services.

  \item \textit{NP-hard Optimization Problem Formulation:} We aim to maximize the two-way secrecy rate of the aerial communication while minimizing excess signal leakage to strengthen system security. Additionally, we seek to reduce AAV energy consumption to maintain energy efficiency. Given the conflicting nature of these goals, we formulate a multi-objective optimization problem (MOP) to jointly optimize these goals. Subsequently, we prove that it is an NP-hard problem.
  


  \item \textit{New Generative Swarm Intelligence Approach:} We propose a novel generative swarm intelligence (GenSI) framework to address the problem. GenSI integrates swarm intelligence with a conditional variational autoencoder (CVAE)-based generative method. In this framework, the CVAE can gather expert solutions from different environment states to generate high-quality initial solutions. Then, swarm intelligence enhanced by leveraging the properties of the formulated MOP will improve these solutions through evolution. 
  
  \item \textit{Simulations and Experiments:} Simulation results demonstrate that the proposed method outperforms various baseline strategies and algorithms while maintaining robustness under phase synchronization error, channel state information (CSI) error, and AAV jitter. Moreover, experimental tests demonstrate that our method can be deployed in limited computing power platforms of AAVs and is beneficial for saving computational resources. Additionally, we find that the initial solution generated by CVAE requires only a few iterations to achieve the same optimization effect that non-generative methods achieve after many iterations starting from their initial solutions.

\end{itemize}

\par The rest of this paper is arranged as follows. Section \ref{sec:related_work} reviews the related work. Section \ref{sec:models_and_preliminaries} presents the models and preliminaries. Section \ref{sec:problem_formulation_and_analysis} formulates the MOP. Section \ref{sec:algorithm} proposes the algorithm. Section \ref{sec:simulation_results_and_analysis} shows the simulation results and Section \ref{sec:conclusion} concludes the paper. 


%
%
\section{Related Work} 
\label{sec:related_work}

\par In this section, we briefly introduce the related work of paradigms, optimization goals, and optimization methods in secure AAV communications to highlight our innovations and contributions.

%
%
\subsection{Paradigms in Secure AAV Communications} 

\par In the literature, multiple paradigms have been proposed to safeguard against eavesdropping, which can generally be categorized into three main types.

\par \textit{Cryptographic Approaches:} Some studies focus on applying cryptographic methods to encrypt AAV communications. For instance, the authors in~\cite{Alladi2020} presented a lightweight mutual authentication scheme using physical unclonable functions for AAV communications, which offers enhanced security features and robustness against common attacks with performance comparisons to existing protocols. Likewise, the authors in~\cite{Aissaoui2023} reviewed the security of communication links for traffic management, compared authentication protocols for AAVs and other constrained systems, and discussed symmetrical alternatives to the AES algorithm along with enhancements for current UTM protocols like ADS-B and RemoteID. Moreover, the authors in~\cite{Hafeez2023} summarized the integration of privacy and security in blockchain-assisted AAV communications and outlined fundamental analyses and critical requirements for building secure and decentralized data systems. Although effective in ensuring confidentiality, cryptographic approaches are often computationally intensive, thus making them less suitable for real-time or resource-constrained AAV systems. Furthermore, they do not inherently address signal-level vulnerabilities, such as eavesdropper collusion, and require additional overhead for key management.

\par \textit{PLS via Trajectory Design and Power Allocation:} Many studies leverage PLS methods in AAV secure communication by exploiting the physical characteristics of wireless channels. On the one hand, trajectory design is another effective approach to handle eavesdroppers. For instance, the authors in~\cite{Wu2023} optimized real-time AAV trajectory for secure integrated sensing and communication, which employed an extended Kalman filter for tracking and predicting ground movements. The authors in~\cite{Zhang2024} designed a multi-AAV mobile edge computing system to enhance secure communication, which proposed a joint dynamic programming and bidding algorithm for AAV trajectory planning. On the other hand, several existing studies considered power allocation-based PLS methods in AAV networks. For example, Maeng \textit{et al.}~\cite{Maeng2022} proposed a linear precoder design for AAVs and derived the optimal power splitting factor. The authors in~\cite{Karmakar2024} introduced FairLearn, an intelligent mechanism in AAV-aided MEC systems that maximizes fairness in secure services by optimizing AAV transmit power and scheduling for task offloading. However, most of these approaches assume non-collaborative eavesdroppers and do not address coordinated attacks, thus limiting their effectiveness in more complex threat environments, such as A2A communications with higher exposure. Thus, this motivates us to investigate a novel PLS to solve the issue of eavesdropper collusion.

\par \textit{PLS Achieved by DCB:} More recent efforts have explored DCB, where multiple AAVs collaborate to direct signals in specific directions, thereby minimizing exposure to potential eavesdroppers. For example, the authors in~\cite{Li2023} enhanced AAV secure communications by using DCB, where AAVs optimize positions and excitation current weights to improve secrecy rates. Moreover, the authors in~\cite{Sun2024} proposed AAV-enabled secure communication with imperfect eavesdropper information by using DCB via an improved multi-objective salp swarm algorithm. In addition, the authors in~\cite{zhang2024multi} developed an AAV swarm-enabled secure surveillance network, which utilizes DCB to enhance data transmission security against eavesdroppers. 
However, these methods are typically designed for individual eavesdroppers and do not consider cases where eavesdroppers collaborate, leaving this an underexplored area.

%
%
\subsection{Optimization Goals of Secure AAV Communications} 

\par In various secure AAV communication paradigms, multiple studies have aimed to balance goals such as security and resource efficiency. The primary optimization objectives are summarized as follows.

\par \textit{Maximizing Secure Indicator:} Several studies focus on enhancing the secure indicator, such as the secrecy rate in AAV communications, which is typically defined as the difference between the transmission rate of the legitimate user and the intercepting rate of the eavesdropper. For instance, the authors in~\cite{Xie2024} developed a multi-AAV cooperative framework to enhance the secrecy rate of AAV communications by utilizing coalitional and Stackelberg game theories, effectively countering eavesdropping attacks. Likewise, the authors in~\cite{Yu2024} investigated the enhancement of secrecy rates in AAV-based ultra-reliable and low latency communications using nonorthogonal multiple access (NOMA), in which optimizing key system parameters to maximize physical layer security through stochastic geometry. Moreover, the authors in~\cite{Wei2024} developed a novel control layer security approach for autonomous systems, which leverages the unobservable cooperative states of AAVs to generate cipher keys, thus enhancing secrecy rates even in environments vulnerable to PLS threats. In addition, the authors in~\cite{Feng2024} developed a strategy for enhancing the secrecy rate in a large-scale multi-tier LEO satellite network through covert communication, which is designed to evade detection by adversarial terrestrial base stations. 

\par \textit{Minimizing Resource Usage of AAV:} Resource usage, particularly energy efficiency, is a critical concern in resource-constrained AAV systems. Numerous studies have optimized the flight trajectory or path planning to minimize energy consumption. Specifically, the authors in~\cite{Li2024} introduced an adversarial multi-agent reinforcement learning scheme for a multi-AAV-assisted MEC system, which targets secure computational offloading and resource allocation against intelligent eavesdroppers and keeps energy efficient. Likewise, the authors in~\cite{Wang2023} developed SEAL, a framework that enhances AAV computation offloading by integrating strategy-proofness, fairness, and privacy preservation through a reverse combinatorial auction mechanism, which significantly improves energy efficiency, protects privacy via off-chain protocols, ensures fairness with on-chain transactions, and reduces operational costs. Moreover, the authors in~\cite{Huang2024} optimized a full-duplex AAV-based communication network, which focuses on energy efficiency by jointly managing 3D trajectory, user scheduling, and power allocation. In addition, the authors in~\cite{Liu2023} developed a game-theoretic routing framework for cellular-connected AAVs, unifying goods delivery and sensing tasks to optimize trajectories, minimize energy consumption, and ensure privacy.  

\par However, existing studies have not established a reasonable secure indicator for the specific case of eavesdropper collusion in A2A communications. This omission creates a gap in addressing the trade-off between maintaining high-security levels and reducing energy consumption in such scenarios.

%
%
\subsection{Optimization Methods for Secure AAV Communications} 

\par The existing methods for solving optimization problems in secure AAV communications can be categorized as follows.

\par \textit{Convex or Non-convex Optimizations:} Some studies employ convex or non-convex optimization techniques to address secure AAV communication problems. For instance, the authors in~\cite{Wu2023} optimized AAV trajectory for secure ISAC using an extended Kalman filtering (EKF)-based method to track and predict the location of a roaming user, and tackled the non-convex trajectory design with an iterative algorithm. Likewise, the authors in~\cite{Zhang2024} employed joint dynamic programming and bidding algorithm alongside successive convex approximation and block coordinate descent for optimizing offloading decisions, resource allocation, and AAV trajectory planning in the secure AAV MEC system. However, these methods are applicable only in scenarios where the problem can be transformed into a convex form. As a result, convex optimization techniques are limited in addressing the inherent non-convexity of the multi-objective problems encountered in secure AAV networks.

\par \textit{Heuristic and Swarm Intelligence Algorithms:} Several studies adopt heuristic approaches, such as genetic algorithms or particle swarm optimization, to find near-optimal solutions for securing AAV communications. For example, the authors in~\cite{Li2024a} tackled a multi-objective optimization problem in secure AAV-assisted IoT networks by proposing an improved swarm intelligence-based algorithm to minimize mission completion time, eavesdropper signal strength, and AAV energy costs. Moreover, the authors in~\cite{Sun2024} addressed the challenges of AAV-enabled secure communications by developing an improved multi-objective salp swarm algorithm to optimize a secure communication multi-objective optimization problem. These algorithms are particularly suited for handling complex, non-convex problems, but they often experience slow convergence and may become trapped in local optima, especially in high-dimensional optimization tasks.

\par \textit{Learning-Based Optimization:} Machine learning techniques, including deep reinforcement learning, have been utilized to dynamically optimize AAV communication parameters. For instance, the authors in~\cite{Yuan2023} developed a DRL-driven framework for AAV-based radio surveillance, and utilized a twin delayed deep deterministic policy gradient (TD3) model to optimize AAV trajectory and RIS configurations for intercepting transmissions without prior channel state information. Moreover, the authors in~\cite{Tham2023} proposed a twin-twin-delayed deep deterministic policy gradient (TTD3) model for real-time optimization of AAV flight trajectory and beamforming in secure RIS-aided AAV communications. In addition, the authors in~\cite{Dong2024} enhanced security in a RIS-enabled AAV communication network by employing the proximal policy optimization (PPO) algorithm, thereby optimizing AAV trajectory, beamforming, and RIS phase shifts for maximizing the average secrecy rate and minimizing secrecy outage duration. While these methods can adapt to changing environments and offer robust performance, they typically require extensive training data and substantial computational resources, which may not be practical in real-time AAV operations.

\par Different from the existing studies, we aim to leverage DCB to address the most severe wiretap scenario involving eavesdropper collusion and seek to propose a scalable and efficient method. This is challenging since we need to balance the secure performance and energy efficiency of the AAV and control complex decision variables in a computation-efficient manner. In the following, we will present the models and preliminaries of the considered system, thereby characterizing the relationships between the decision variables with secure performance and energy efficiency of the system.

%
%
\section{Models and Preliminaries} 
\label{sec:models_and_preliminaries}

\par In this section, we first present the system overview. Then, we detail the considered models, including the network model and AAV energy cost models, to characterize the objectives, decision variables, and environmental factors. The main notations are presented in Table \ref{tab:notations}.


\begin{table}[]
\centering
\caption{Main notations}
\label{tab:notations}
\begin{tabularx}{3.5in}{p{1.5cm}p{6.5cm}}
\toprule
\textbf{Symbol}         & \textbf{Definition}                                                     \\ \midrule
$\mathcal{A}$ & The archive of the proposed GenSI framework \\
$AF_i$ & The array factor of the $i$th AVAA \\
$\boldsymbol{c}$ & The conditional information (i.e., environment factors) of the CVAE model \\
$C_{KE}$ & The minimum two-way known secrecy capacity of the system\\
$\mathcal{E}_K$ & The sets of the known eavesdroppers \\
$\mathcal{E}_U$ & The sets of the unknown eavesdroppers \\
$E_{i,j}$ & The energy cost of $j$th AAV of $i$th AVAA for performing the communication \\
$f_{SLL_i}$ &  The maximum SLL of $i$th AVAA \\
 $G^{A2A}_i$ & The antenna gain of the $i$th AVAA \\
$N_{U}$ & The number of the AAVs within a swarm \\
$N_{E}$ & The number of the AAVs within a swarm \\
$\mathcal{P}$ & The population of the proposed GenSI framework \\
$\boldsymbol{P}$ & The positions of the AAVs of two AVAAs \\
$P^{\text{LoS}}_{i,k}$ & The probability of $i$th AVAA and eavesdropper $k$\\
$\boldsymbol{\Omega}$ & The excitation current weights of the AAVs of two AVAAs \\
$\mathcal{U}_1$, $\mathcal{U}_2$ & The sets of the rotary-wing AAVs in the first and second swarms, respectively \\
$\boldsymbol{u}$ & The select AAV receivers of two AVAAs \\
$\boldsymbol{X}$ & The solution to the formulated MOP \\
$\boldsymbol{z}$ & The latent variable of the CVAE model \\

\bottomrule
\end{tabularx}
\end{table}

%
\subsection{System Overview} 

\par
\begin{figure}
  \centering
  \includegraphics[width=3.5in]{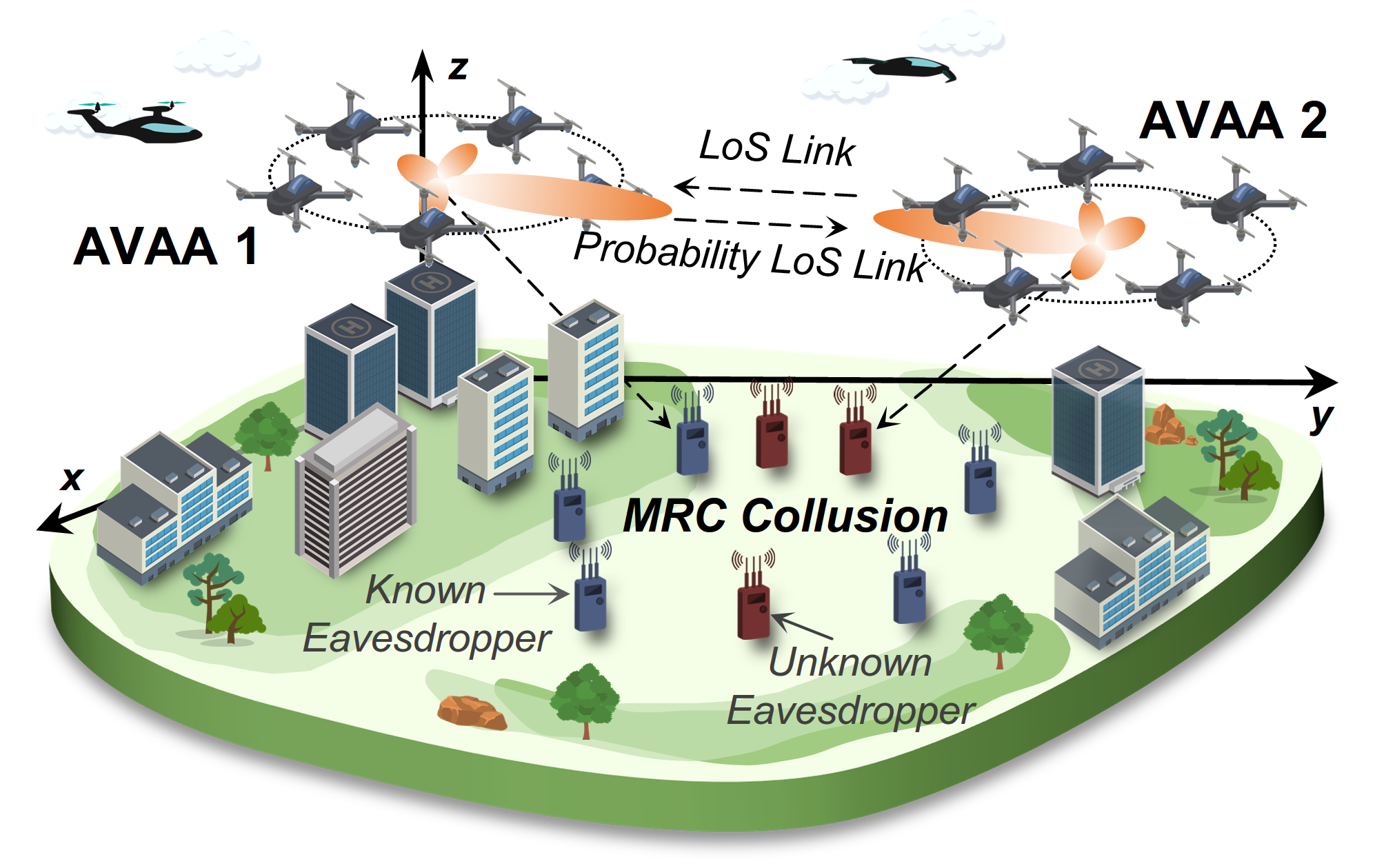}
  \caption{A two-way DCB-enabled aerial communication system under the known and unknown eavesdropper collusion in LAE scenarios.}
  \label{fig:network-model}
\end{figure}

 As illustrated in Fig.~\ref{fig:network-model}, we consider a long-range two-way aerial communication system under eavesdropper collusion in LAE scenarios. Specifically, the system comprises two AAV swarms denoted as $\{ \mathcal{U}_i | i \in 1, 2\}$ deployed in a network-deficient area for emergency assistance, wildlife surveillance, military operations, etc. Note that we can easily extend two AAV swarm systems to multiple by introducing routing and networking protocols, such as greedy perimeter stateless routing, opportunistic routing, mobility prediction-based geographic routing, etc~\cite{Javed2024}. We focus on two AAV swarms for the sake of simplicity and easy access to insights in this work. Accordingly, a set of randomly distributed AAVs denoted as $\mathcal{U}_1 \triangleq \{j | 1, 2, \dots, N_{U}\}$ are in a tight area denoted as $A_{U1}$. Moreover, another AAV swarm denoted as $\mathcal{U}_2 \triangleq \{j | 1, 2, \dots, N_{U}\}$ is dispatched in tight area $A_{U2}$ which is quite far and no overlapping from $A_{U1}$. These two areas are compact in scale but geographically well-separated, resulting in critical communication challenges between the two AAV swarms.

\par In both the areas and the surrounding ground, there are several eavesdroppers denoted as $\mathcal{E} \triangleq \{k | 1, 2, \dots, N_{E}\}$ that are randomly placed. We consider that some eavesdroppers can be detected~\cite{Zhang2019} while others are undetectable but are potential threats. Accordingly, the eavesdroppers in $\mathcal{E}$ can be divided into known eavesdroppers and unknown eavesdroppers as $\mathcal{E}_K$ and $\mathcal{E}_U$, respectively. Moreover, we consider that each AAV of $\mathcal{U}_1$ and $\mathcal{U}_2$ is equipped with a single omni-directional antenna, and all the eavesdroppers collude.
 
\par The system operates as follows. Consider a particular case that the AAVs in $\mathcal{U}_1$ and $\mathcal{U}_2$ need to exchange emergency information and data. Due to the lack of terrestrial access points, these two AAV swarms will construct two AVAAs (\textit{i.e.}, AVAA 1 and AVAA 2) simultaneously. We consider that the AAVs within the same virtual antenna array are synchronized regarding the carrier frequency, time, and initial phase by using the methods in~\cite{Mohanti2022}. Moreover, the data sharing among AAVs of each virtual antenna array is achieved by using the method in~\cite{Feng2013}. In addition, we consider that the AAVs obtain the quantization version of CSI via method in~\cite{Ahmad2022}. Following this, they will select a suitable receiver from a different AAV swarm as the receiver and then achieve two-way aerial communication. Note that the overhead of synchronization and information sharing in terms of time and energy is relatively small. This is because the relatively compact distribution of AAVs can lead to an effective inner-swarm communication condition. As such, according to~\cite{Abdelkader2019, Maina2022, Ouassal2021}, the overhead is determined by the specific synchronizing and controlling protocols, and only accounts for a small portion of overall communications, and thus can be ignored.

\par Without loss of generality, we consider a 3D Cartesian coordinate system. Following this, the positions of the $j$th AAV in $i$th swarm $\mathcal{U}_i$ and the $k$th eavesdropper are indicated as $(x^{U}_{i,j}, y^{U}_{i,j}, z^{U}_{i,j})$ and $(x^{E}_{k}, y^{E}_{k}, 0)$, respectively. In what follows, we present the models within the system including the network model and AAV energy cost model.

%

\subsection{Network Model}

\par In this subsection, we present the network model of the considered system, which primarily focuses on far-field communication scenarios due to the long-range communication requirements of the AAVs. 

\subsubsection{Virtual Antenna Array Model} 

\par The electromagnetic waves transmitted by AAV antennas within an AVAA undergo superposition and interference, resulting in a beam pattern characterized by a mainlobe and sidelobes. Mathematically, we utilize array factor to quantify signal intensities in various directions. Let $(x^{U}_{i,j}, y^{U}_{i,j}, z^{U}_{i,j})$ and $\omega_{i,j}$ represent the \textit{3D coordinates} and \textit{excitation current weight} of the $j$th AAV in $\mathcal{U}_i$, respectively. Then, the array factor of the $i$th AVAA is expressed as follows:
\begin{equation}
   \label{eq:AF}
   \begin{aligned}
      A&F_i(\theta, \phi)= \\
      &\sum \limits _{j = 1}^{N_{U}} \omega_{i,j} e^{j^{u} \left [ \frac{2\pi}{\lambda} \left ({{{ x^{U}_{i,j} \sin \theta \cos \phi + {y^{U}_{i,j}}\sin \theta \sin \phi + {z^{U}_{i,j}}}\cos \theta } }\right) + \Phi_{i,j}\right]},
   \end{aligned}
\end{equation}
\noindent where $\theta$ and $\phi$ which range $[0, \pi]$ and $[-\pi, \pi]$ are the elevation and azimuth angles from the center of $\mathcal{U}_i$ ($x^c_i$, $y^c_i$, $z^c_i$) to any receiver, respectively. Other parameters shown in Eq.~\eqref{eq:AF} are related to communications, \textit{e.g.}, $j^{u}$ is the imaginary unit and $\Phi_{i,j}$ is the initial phase of $j$th array element in $i$th AVAA. Note that we considered perfect phase alignment among AAV antennas to avoid random perturbations that could interfere with the subsequent optimization and solution processes. In this case, the impact of imperfect phase alignment on system performance will be evaluated in Section~\ref{sec:simulation_results_and_analysis}.

%
\subsubsection{A2A Transmission Model}

\par Due to the high altitudes of AAVs and usage of DCB, the A2A transmission should follow a LoS channel condition~\cite{Mozaffari2019}. Let $d^{A2A}_i$ denote the distance between the center of $i$th AVAA and the corresponding receiver, the transmission rate is given by
\begin{equation}
  \label{eq:R_A2A}
  \begin{aligned}
    R^{A2A}_i=B \log _{2}(1+\frac{P^{t}_i K_{0} G_{i}^{A2A} {d^{A2A}_i}^{-\alpha}}{\sigma^{2}}).
  \end{aligned}
\end{equation}
\noindent Other parameters in Eq.~\eqref{eq:R_A2A} are related to communications. Specifically, $B$ is the bandwidth, $P^t_i$ is the total transmission power of the $i$th AVAA, $K_{0}$ is the constant path loss coefficient, and $\sigma^{2}$ is the noise power. Another key parameter in Eq.~\eqref{eq:R_A2A} is the antenna gain $G_{i}^{A2A}$. Let $(\theta_i, \phi_i)$ denote the direction from the center of $i$th AVAA to the receiver, $G_{i}^{A2A}$ is expressed as 
\begin{equation}
  \label{eq:gain}
  \begin{aligned}G_{i}^{A2A}=\frac{4\pi\left|AF_i\left(\theta_{i}, \phi_{i}\right)\right|^{2} w\left(\theta_{i}, \phi_{i}\right)^{2}}{\int_{0}^{2 \pi} \int_{0}^{\pi}|AF_i(\theta, \phi)|^{2} w(\theta, \phi)^{2} \sin \theta \mathrm{d} \theta \mathrm{d} \phi} \eta ,
  \end{aligned}
\end{equation}
\noindent where $w{(\theta,\phi)}$ is the magnitude of the far-field beam pattern of each AAV, $\eta \in [0, 1]$ is the antenna array efficiency~\cite{Mozaffari2019}.

%
\subsubsection{Eavesdropper Collusion Model} 

\par The ground-placed eavesdroppers suffer a probability LoS channel condition. Let $\theta_{i, k}$ denote the elevation angle between the center of $i$th AVAA and eavesdropper $k$, then the LoS probability is~\cite{Faraci2019}
\begin{equation}
  \label{eq:LoS_P}
  \begin{aligned}
    P^{\mathrm{LoS}}_{i,k}=\left(1+b_1 e^{-b_2(\theta_{i, k}-b_1)} \right)^{-1},
  \end{aligned}
\end{equation}
\noindent where $b_{1}$ and $b_{2}$ represent environment-dependent constants. As the elevation angle $\theta_{i,k}$ is directly determined by the eavesdropper location $(x^{E}_{k}, y^{E}_{k}, 0)$, these spatial coordinates serve as key environmental parameters impacting security performance. Following this, the non-line-of-sight (NLoS) probability is given by $P^{\mathrm{NLoS}}_{i,k}=1-P^{\mathrm{LoS}}_{i,k}$. Accordingly, let $d_{i,k}$ and $G_{i,k}$ denote the distance and antenna gain between the center of $i$th AVAA and the eavesdropper $k$, respectively, the corresponding SNR is expressed as 
\begin{equation}
  \begin{aligned}
    \gamma_{i,k}= \frac{P^t_i K_0 G_{i,k} d_{i,k}^{-\alpha}[P^{\mathrm{LoS}}_{i,k} \mu_{\mathrm{LoS}}+P^{\mathrm{NLoS}}_{i,k} \mu_{\mathrm{NLoS}}]^{-1}} {\sigma^2},
  \end{aligned}
\end{equation}
\noindent where $\mu_{\mathrm{LoS}}$ and $\mu_{\mathrm{NLoS}}$ are the attenuation factors of LoS and NLoS links, respectively. Note that the exclusion of short-term fading in this model is based on two key considerations. First, high-altitude air-to-air channels primarily exhibit LoS propagation, which minimizes the impact of multipath effects. Second, we aim to deploy the AAV-enabled virtual antenna array to achieve better performance over an extended post-deployment period. In this case, we are more interested in the expected average performance metrics over a relatively long time period rather than dynamic instantaneous performance. As such, short-term fading may add more dynamic and affect the expected average performance.

\par The eavesdroppers employ maximum ratio combining (MRC) based on signal detection~\cite{Dung2021}. This is because MRC-based collusion represents the worst wiretap case and provides a foundation for analyzing other collusion patterns. In particular, each eavesdropper can be regarded as an antenna in a multi-antenna system. Then, the system weights and combines the received signals from colluded eavesdroppers via the MRC diversity-combining technique~\cite{Dung2021}. In this case, following the optimal weight assignment across different branches, the combined output SNR from the $i$th AVAA is given by
\begin{equation}
  \label{eq:snr_all}
  \begin{aligned}
    {\gamma_{\Sigma}}_i= \sum_{k=1}^{N_{E}} \gamma_{i,k}.
  \end{aligned}
\end{equation}

\noindent Thus, the achievable rate of colluding eavesdroppers from $i$th AVAA is given by $R_{i}^{E}= B \log_2 (1+ {\gamma_{\Sigma}}_i)$. Using $R^{A2A}_i$ and $R_{i}^{E}$, we can express the \textit{minimum two-way achievable secrecy capacity} of the two-way communication as
\begin{equation}
  \label{eq:ce}
  \begin{aligned}
    C_E=\min_{i \in \{1, 2\}, k \in \mathcal{E}} \{R^{A2A}_i- R_{i}^{E}\}.
  \end{aligned}
\end{equation}

\noindent As can be seen, the defined secure transmission performance considers the minimum capacity between two-way communications, which means that improving $C_E$ can enhance the secure performance in both AVAA 1 to AVAA 2 and AVAA 2 to AVAA 1. 

\par Nevertheless, when we formulate the optimization problem, only the information of known eavesdroppers is available. Thus, 
we introduce the concept of the \textit{minimum two-way known secrecy capacity} as follows:
\begin{equation}
  \label{eq:cke}
  \begin{aligned}
    C_{KE}=\min_{i \in \{1, 2\}, k \in \mathcal{E}_K} \{R^{A2A}_i- R_{i}^{E}\}.
  \end{aligned}
\end{equation}

\noindent It can be seen from Eqs.~\eqref{eq:LoS_P}-\eqref{eq:cke} that both the minimum two-way achievable secrecy capacity and minimum two-way known secrecy capacity are determined by the array factors of the AVAAs and affected by eavesdropper positions. In other words, by considering the eavesdropper locations, we can strategically adjust the signal distribution of the AVAAs to enhance security performance.

\subsection{AAV Energy Cost Model}

\par We consider the typical rotary-wing AAV, where the propulsion energy cost dominates the total energy expenditure while other components can be negligible~\cite{Sun2021}. Let $v$ be the flying speed of the AAV, the energy cost of the AAV for flying in the horizontal plane is given by
\begin{equation}
   \label{eq:energy-2d}
   \begin{aligned}
      P(v)=&P_{B}(1+\frac{3v^2}{v_{tip}^2})+P_{I}(\sqrt{1+\frac{v^4}{4v_0^4}}-\frac{v^2}{2v_0^2})^{1/2}+\\&\frac{1}{2}d_0\rho sAv^3,
   \end{aligned}
\end{equation}
\noindent where $P_{B}$, $P_{I}$, $v_{tip}$, $v_{0}$, $d_{0}$, $s$, $\rho$, and $A$ are blade profile constant, induced powers constant, tip speed of the rotor blade, mean rotor induced velocity in hovering, fuselage drag ratio, rotor solidity, air density, and rotor disc area, respectively.

\par According to Eq.~\eqref{eq:energy-2d}, the propulsion energy cost of AAV can be extended into the 3D form, \textit{i.e.}, 
\begin{equation}
  \label{eq:energy-3d}
   \begin{aligned}
     E(T) \approx &\int_0^TP( v(t))dt+{\frac12m_{D}(v{(T)}^2- v{(0)}^2)}+\\&{m_{D}g(h(T)-h(0))},
   \end{aligned}
\end{equation}
\noindent where $v(t)$ is the instantaneous AAV speed at time $t$, $T$ represents the end time of the flight, $m_{D}$ is the aircraft mass of an AAV, and $g$ is the gravitational acceleration. 

\par Note that we introduce the trajectory design and speed control schemes in~\cite{Sun2021}. Based on the schemes and Eq.~\eqref{eq:energy-3d}, the energy consumption of the AAVs is determined by initial positions and the positions of the AAVs for performing VAA. Clearly, the initial positions of the AAVs are the environmental factors, while the positions of the AAVs for performing VAA need to be optimized and are the decision variables. We also note that we do not include the energy consumption of synchronization and data sharing in the energy model. As mentioned previously,  the energy overhead of these preliminary steps can be relatively small and ignored compared to the overall system operation. Moreover, these energy overheads are primarily determined by the synchronizing and controlling protocols rather than the AAV-related decision variables. Therefore, excluding these energy overheads from our energy model does not significantly impact the optimal solution.

%
\section{Problem Formulation and Analysis} 
\label{sec:problem_formulation_and_analysis}

\par In this section, we formulate a secure and energy-efficient two-way aerial communication problem. Firstly, we state the main idea of the optimization problem. Secondly, we present the decision variables and optimization objectives. Finally, we formulate an MOP and prove that it is an NP-hard problem.

\par The considered system for LAE concerns two goals, \textit{i.e.}, improving the achievable secrecy capacity and reducing the energy cost for fine-tuning the positions of AAVs. \textit{On the one hand}, since unknown eavesdropper information is unavailable, we cannot directly optimize the minimum two-way secrecy capacity in Eq.~\eqref{eq:ce}. As such, we utilize two measures to optimize the achievable secrecy capacity simultaneously. Firstly, we aim to maximize the known secrecy capacity as given in Eq.~\eqref{eq:cke} to reduce the signal qualities obtained by known eavesdroppers. Secondly, we can minimize all the signals except the target direction, thereby avoiding potential unknown eavesdroppers. These two measures can be achieved by controlling the array factor (which is determined by the \textit{3D positions} and \textit{excitation current weights}) and \textit{aerial receiver} selection. \textit{On the other hand}, we need to fine-tune AAV positions to enhance DCB performance, which will result in extra energy costs. Thus, the \textit{position changes} of AAVs should be minimized by considering the energy efficiency. 

\par Based on the above analyses, the following decision variables need to be determined jointly: \textit{(i)} $\boldsymbol{P} = \left\{(x_{i,j}^U, y_{i,j}^U, z_{i,j}^U) | i \in \{ 1, 2 \}, j \in \mathcal{U}_i \right\}$, a matrix consisting of continuous variables denotes the 3D positions of AAVs in AVAA1 and AVAA2 for performing DCB. \textit{(ii)} $\boldsymbol{\Omega} = \left\{ \omega_{i,j} | i \in \{ 1, 2 \}, j \in \mathcal{U}_i \right\} $, a matrix consisting of continuous variables denotes the excitation current weights of AAVs in AVAA1 and AVAA2 for performing DCB. \textit{(iii)} $\boldsymbol{u} = \left\{ u_i | i \in \{1, 2\} , u_i \in \mathcal{U}_i \right\} $, a vector consisting of discrete variables represents the IDs of selected aerial receivers of AVAA 1 and AVAA 2. 

\par As such, we consider the following objectives to achieve DCB-enabled secure and energy-efficient two-way communication in LAE.

\par \textit{Optimization Objective 1}: The first objective is to maximize the minimum two-way known secrecy capacity of the system to avoid known eavesdroppers. To this end, we need to jointly optimize the $\boldsymbol{P}$, $\boldsymbol{\Omega}$, and $\boldsymbol{u}$. Thus, our first objective is expressed as follows: 
\begin{equation}
  \label{eq: objecitve1}
  \begin{aligned}
    f_{1}(\boldsymbol{P}, \boldsymbol{\Omega}, \boldsymbol{u})= C_{KE}.
  \end{aligned}
\end{equation}

\par \textit{Optimization Objective 2}: The second objective is to minimize the signal densities of the AVAAs in all directions to avoid unknown eavesdroppers. Specifically, we adopt the \textit{maximum sidelobe level (SLL)} for evaluating the signal strength except for the targeted direction of an AVAA. Let $(\boldsymbol{\theta}^{SLL}_i, \boldsymbol{\phi}^{SLL}_i)$ denote the direction set except targeted direction, and then the maximum SLL of $i$th AVAA is given by
\begin{equation}
  \begin{aligned}
    f_{SLL_i} = \frac{ \max|AF_i(\boldsymbol{\theta}^{SLL}_i, \boldsymbol{\phi}^{SLL}_i)|}{AF_i(\theta_i, \phi_i)}.
  \end{aligned}
\end{equation}
\noindent Based on this, we seek to minimize the maximum term among the maximum SLLs of AVAAs. To this end, we jointly optimize $\boldsymbol{P}$ and $\boldsymbol{\Omega}$. As such, our second objective can be expressed as
\begin{equation}
  \label{eq: objecitve2}
  \begin{aligned}
    f_{2}(\boldsymbol{P}, \boldsymbol{\Omega})= \max_{i \in \{1, 2\}} \{ f_{SLL_i} \}. 
  \end{aligned}
\end{equation}

\noindent Accordingly, we can minimize Eq.~\eqref{eq: objecitve2} to optimize the maximum SLLs of AVAA 1 and AVAA 2 simultaneously. 

\par \textit{Optimization Objective 3}: The third objective is to minimize the energy costs of all AAVs when fine-tuning their positions. To this end, we optimize the $\boldsymbol{P}$ to achieve this goal, and our third objective is expressed as follows: 
\begin{equation}
  \label{eq: objecitve3}
  \begin{aligned}
    f_{3}(\boldsymbol{P})= \sum_{i \in \{1, 2\}} \sum_{j \in \mathcal{U}_i} E_{i,j},
  \end{aligned}
\end{equation}
\noindent where $E_{i,j}$ is the energy cost of $j$th AAV of $i$th AVAA for moving to the DCB position. Note that we calculate $E_{i,j}$ according to $\boldsymbol{P}$ and the original position of AAVs $\boldsymbol{P}^{r}$ by using the method in~\cite{Sun2021}.

\par By considering the three conflicting objectives, the optimization problem can be given in an MOP formulation as 
\begin{subequations}
  \label{eq:formulation}
  \begin{align}
    {\underset{\boldsymbol{X} = \{\boldsymbol{P}, \boldsymbol{\Omega}, \boldsymbol{u}\} }{\text{min}}} \ & F=\{-f_{1}, f_{2}, f_{3}\},\\
    \text{s.t.} \quad \quad
    & (x_{i,j}^U, y_{i,j}^U, z_{i,j}^U) \in \mathbb{R}^{3 \times 1}_i , \forall i \in \{1, 2\}, \forall j \in \mathcal{U}_i, \label{eq:const1}\\
    & 0 \leqslant \omega_{i, j} \leqslant  1, \forall i \in \{1, 2\}, \forall j \in \mathcal{U}_i \label{eq:const2},\\
    & u_i \in \mathcal{U}_{\{1,2\}-i}, \forall i \in \{1, 2\} , u_i \in \mathcal{U}_i \label{eq:const3}, \\
    & d_{(j_1, j_2)}^i \geq d_{min}, \forall i \in \{1, 2\}, \forall j_1, j_2 \in \mathcal{U}_i \label{eq:const4},
  \end{align}
\end{subequations}

\noindent where $\boldsymbol{X}$ is the decision variable set of the problem. Then, $\mathbb{R}^{3 \times 1}_i$ is the reachable area of the $i$th AVAA. Moreover, $\omega_{i, j}$ indicates the ranges of the excitation current weights. In particular, $\omega_{i, j}=0$ means the antenna array of the AAV is switched off, while $\omega_{i, j}=1$ indicates that the antenna transmits signals in the maximum transmission power. In addition, constraint \eqref{eq:const3} shows that the AVAA must select a receiver from a different AAV swarm. Furthermore, constraint \eqref{eq:const4} presents a hard condition, \textit{i.e.}, minimum separation distance, to avoid the collision between any AAVs in all AVAAs. Note that the formulated optimization problem shown in Eq.~\eqref{eq:formulation} can be simplified as a nonlinear multi-dimensional knapsack problem. Thus, the optimization problem is NP-hard. In what follows, we provide the proof. 

\par Specifically, assume that the optimization problem is simplified by only considering minimizing the SLL goal in one-way communication with the determined $\boldsymbol{P}$ and $\boldsymbol{u}$. Then, the decision variables of the simplified problem are $\boldsymbol{X}'= [\omega_1, \omega_2,..., \omega_{N_{U}}]$, where $0 \le \omega_j \le 1$ and $j \in \mathcal{U}_1$. We further simplified $\boldsymbol{X}'$ as the discrete solution, \textit{i.e.}, $\omega_j \in \mathcal{S}=\{0,1\}$, and we define a new function $g(\boldsymbol{X}')= \sum_{1}^{N_{U}}\omega_j$. Clearly, $g(\boldsymbol{X}')<N_{U}$, wherein $N_{U}$ is a constant. Thus, the simplified problem is given by
	\begin{subequations}
	  \label{S-SECMOP}
	  \begin{align}
	    {\underset{\boldsymbol{X}'}{\text{min}}}  \quad  & F=f_{SLL_1}\\
	    \text{s.t.} \quad & g(\boldsymbol{X}')<N_{U},\\
	    & \omega_{j} \in S, \forall j \in \mathcal{U}_1.
	  \end{align}
	\end{subequations}
	\noindent As such, the simplified problem is in the form of a nonlinear multi-dimensional knapsack problem \cite{Goos2020}, and it has a similar computational complexity to the 0-1 knapsack problem which is NP-hard. Thus, our problem in Eq.~\eqref{eq:formulation} is NP-hard since it is much more complex than the simplified problem.

\par As such, due to the computational complexity, the problem cannot be solved in polynomial time. Moreover, the intricate and nonlinear relationships between the decision variables and the optimization objectives make it challenging to devise a powerful approximation algorithm or convex optimization method~\cite{Li2024a}. Artificial intelligence methods are promising to solve such complex problems efficiently, \textit{e.g.}, deep reinforcement learning and swarm intelligence algorithms can solve sequence decision-making and static deployment problems, respectively. Considering that our problem is static, reinforcement learning, which excels in sequential decision-making and dynamic environments, may not be suitable. In this case, swarm intelligence algorithms are promising for solving such static and complex optimization problems due to their global search capabilities and robustness. 

\par However, traditional swarm intelligence algorithms typically start their search in a nearly random manner each time. This means that for every new problem instance (e.g., the instances with different environmental factors), the algorithm must explore the solution space anew, which can be time-consuming and computationally intensive. In the considered problem, different instances may share underlying patterns and experiences that can be learned and utilized to facilitate the optimization process. For example, optimal AAV configurations or excitation current weights in previous scenarios might exhibit certain characteristics that are also beneficial in new scenarios. If we can capture these patterns, we may provide better initial solutions to the swarm intelligence algorithm, thereby enhancing its efficiency and convergence speed.

\par To leverage this potential, we aim to propose a novel GenSI framework that synergistically combines swarm intelligence algorithms with generative artificial intelligence in the following.

%
\section{Algorithm Design} 
\label{sec:algorithm}

\par In this section, we present the GenSI framework, which consists of two main phases, \textit{i.e.}, initialization and evolution. We first present the structure of the GenSI framework, and then detail the initialization and evolution phases of the GenSI framework.

\begin{figure*}
  \centering
  \includegraphics[width=1 \textwidth]{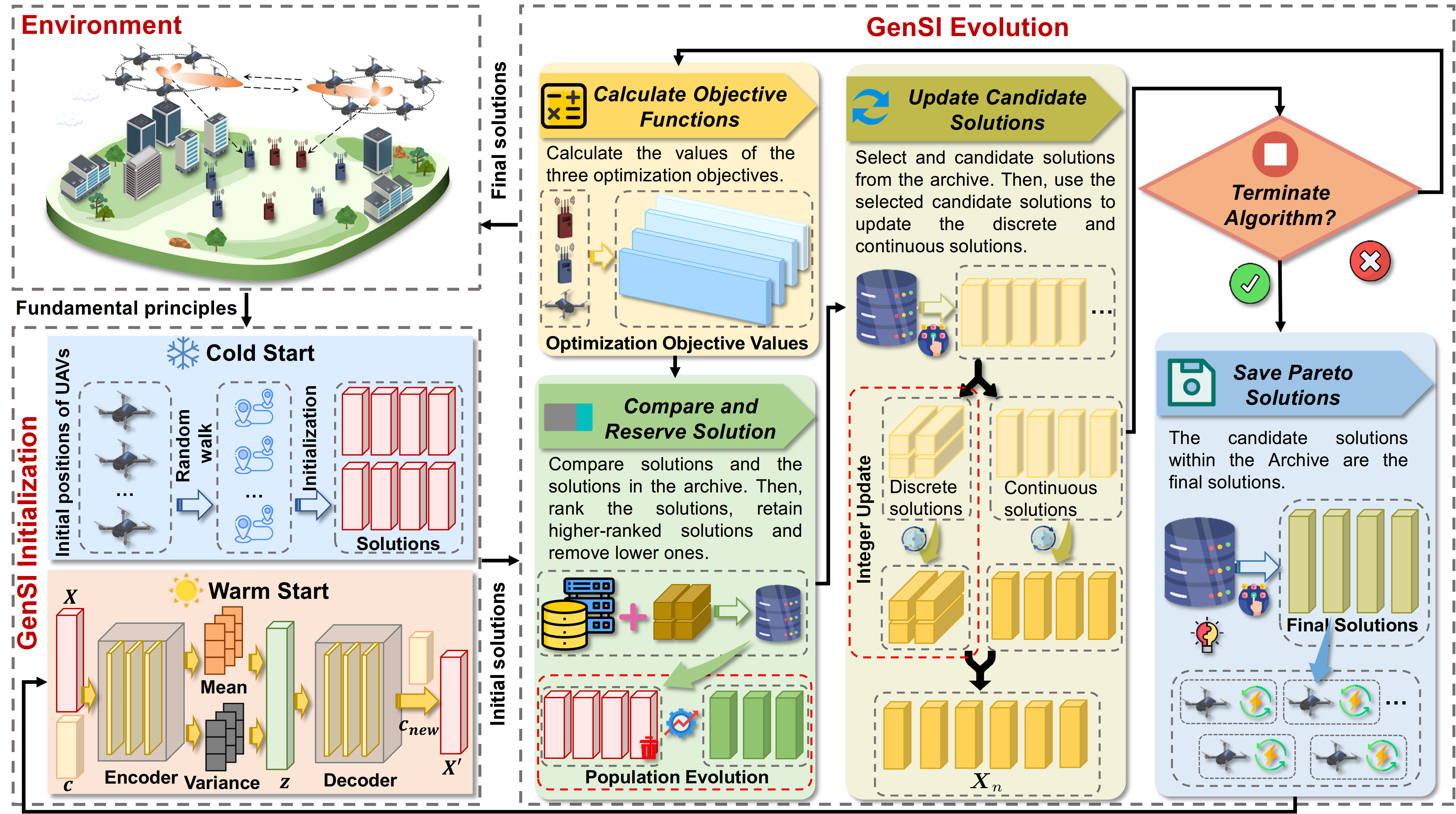}
  \caption{The structure of the GenSI framework.}
  \label{fig:Framework}
\end{figure*}

%
\subsection{Structure of GenSI Framework}
\label{ssec:gensi_framework}

\par As shown in Fig.~\ref{fig:Framework}, the GenSI framework works in two phases which are an initialization phase and an evolution phase. Specifically, in the initialization phase, the GenSI framework aims to generate an initial population of candidate solutions. Depending on the scenario, the framework either employs a random walk-based initialization if no trained model is available (cold start) or uses a powerful generative artificial intelligence model, namely, CVAE, to generate high-quality initial solutions (warm start). This initial population serves as a starting point for the optimization process and significantly impacts the efficiency and quality of the final solution.

\par Moreover, in the evolution phase, the GenSI framework utilizes a multi-objective optimization method to iteratively evolve the candidate solutions. This process involves selecting, evaluating, and updating the population, aiming to improve the solutions based on defined optimization objectives. The evolution phase leverages the swarm intelligence principles to guide the search more effectively, which provides an adaptive mechanism to navigate the solution space efficiently.

\par In essence, the initialization phase aims to provide a well-informed starting point that can lead to better convergence, while the evolution phase seeks to refine and optimize the candidate solutions through iterative improvement, thus leveraging both learned patterns from generative models and swarm intelligence. In the following, we present these two phases in detail.

%
\subsection{GenSI Initialization}
\label{ssec:gensi_initialization}

\par During the initialization phase, the GenSI framework aims to generate a set of candidate solutions for subsequent evolution and improvement. To this end, the GenSI framework needs to decide between a cold start and a warm start based on the availability of a trained CVAE model.

%
%
\subsubsection{Cold Start with Random Walk-based Initialization}
\label{sssec:cold_start_initialization}

\par In a cold start scenario, where no trained CVAE is available, random walk-based initialization is used to generate the initial candidate solutions. This method ensures that the initial solutions are adequately diverse, facilitating the exploration of the solution space in subsequent iterations. To ensure the quality of these candidate solutions, we can summarize some fundamental principles and add them to the initialization phase.

\par Specifically, the region near $\boldsymbol{P}^r$ holds particular significance, since we are more likely to find positions of AAVs with low energy consumption and relatively high secure performance. According to Eq.~\eqref{eq: objecitve3}, the third optimization objective refers to the energy consumption of AAVs for flying from their original positions $\boldsymbol{P}^r$ to the optimized positions. 
Consequently, proximity to $\boldsymbol{P}^r$ results in lower values for this objective function. Meanwhile, the first and second objectives (in Eqs.~\eqref{eq: objecitve1} and~\eqref{eq: objecitve2}) depend on the AVAA array factor, which is controlled by the relative positions of AAVs. Thus, even if AAVs fly around $\boldsymbol{P}^r$, they can still attain the optimal relative positions, thereby achieving a relatively high level of secure performance.

\par Motivated by this, we propose a random walk-based initialization method that employs the information of $\boldsymbol{P}^r$ and a random walk approach to design the initial state of $\boldsymbol{P}$ part of each candidate solution $\boldsymbol{X}$. \textit{First}, we generate the $N$ vectors drifted from $\boldsymbol{P}^r$ by using random walk, which is given by
\begin{equation}\label{eq:random_walk}
\begin{aligned}
	&\boldsymbol{P}^*= [ \boldsymbol{P}^*_1, \boldsymbol{P}^*_2,...,\boldsymbol{P}^*_N ] \\&=\left[0, f_c\left(2 r\left(1\right)-1\right), f_c \left(2 r\left(2\right)-1\right), \ldots, f_c\left(2 r\left(N\right)-1\right)\right],
\end{aligned}
\end{equation}

\noindent where $f_c$ calculates the cumulative sum, and $r(n)$ is a stochastic function which is given by 
\begin{equation}
	\label{eq:sto}
	r(n)=\left\{\begin{array}{ll}
1 & \text { if } rand >0.5 \\
0 & \text { if } rand \leqslant 0.5
\end{array}\right.,
\end{equation}
\noindent where $rand$ is a random number generated with uniform distribution in the interval of $[0, 1]$. 

\par \textit{Second}, we use these vectors and random number generator to initialize the $n$th candidate solution of the population $\boldsymbol{X}_{n}^{init}$ as follows:
\begin{equation}\label{eq:generate}
\begin{aligned}
	&\boldsymbol{X}_{n}^{init}= [ \boldsymbol{P}^*_n, \boldsymbol{\Omega}_{rand}, \boldsymbol{u}_{rand}],
\end{aligned}
\end{equation}
\noindent where $\boldsymbol{\Omega}_{rand}$ is a random vector, an element of which is a random number generated with uniform distribution in the interval of $[0, 1]$. Moreover, $\boldsymbol{u}_{rand}$ is a random integer vector with two elements, and each element can be calculated as $round\left(rand*N_{U}\right)$.

\par By using this method, GenSI can generate more informative initial candidate solutions even if in a cold start case where no trained CVAE is available, thereby ensuring the availability of cold deployment of the GenSI.

%
%
\subsubsection{Warm Start with CVAE-Based Initialization}
\label{sssec:warm_start_initialization}

\par In a warm start scenario, the generative artificial intelligence model, CVAE, is employed to generate the initial candidate solutions. In particular, the CVAE is trained by using the archive of Pareto-optimal solutions obtained from previous optimization runs, where the environment factors are used as conditional information. This enables the CVAE to learn the patterns and relationships between environmental conditions and high-quality candidate solutions, allowing it to generate promising initial solutions for new optimization instances. We begin by presenting the principles of CVAE and then introduce the details of how it is utilized for initialization.

\par Specifically, the CVAE consists of an encoder and a decoder. The encoder maps the input solutions to a latent space, while the decoder reconstructs the solutions from the latent space, conditioned on the environment factors. Mathematically, the encoder aims to approximate the posterior distribution $q_{\phi}(\boldsymbol{z} | \boldsymbol{X}, \boldsymbol{c})$, where $\boldsymbol{X}$ represents the decision variables (i.e., the solutions), $\boldsymbol{c}$ denotes the conditional information (i.e., environment factors), and $\boldsymbol{z}$ is the latent variable. The decoder, on the other hand, attempts to reconstruct the solution $\boldsymbol{X}$ given the latent representation $\boldsymbol{z}$ and the conditional information $\boldsymbol{c}$, modeled as $p_{\theta}(\boldsymbol{X} | \boldsymbol{z}, \boldsymbol{c})$.

\par The objective of the CVAE is to minimize the following loss function, which consists of two main components:
\begin{equation}
\begin{aligned}
    \mathcal{L}& (\phi, \theta; \boldsymbol{X}, \boldsymbol{c}) = \\
    &\mathbb{E}_{q_{\phi}(\boldsymbol{z} | \boldsymbol{X}, \boldsymbol{c})} \left[ -\log p_{\theta}(\boldsymbol{X} | \boldsymbol{z}, \boldsymbol{c}) \right] + D_{KL} \left( q_{\phi}(\boldsymbol{z} | \boldsymbol{X}, \boldsymbol{c}) \parallel p(\boldsymbol{z}) \right),
\end{aligned}
\end{equation}
\noindent where the first term represents the reconstruction loss, which ensures that the generated solution $\boldsymbol{X}$ matches the original solution, and the second term is the Kullback-Leibler (KL) divergence, which regularizes the latent variable $\boldsymbol{z}$ to follow a prior distribution $p(\boldsymbol{z})$, typically a standard normal distribution.

\par In this work, the environment factors $\boldsymbol{c}$ include the initial positions of AAVs, the positions of eavesdroppers, and other relevant environmental parameters that influence the optimization process. The decision variables $\boldsymbol{X}$ are be optimized, such as AAV positions, beamforming weights, and receiver selections. As shown in Fig.~\ref{fig:Framework}, the encoder compresses the relationship between $\boldsymbol{X}$ and $\boldsymbol{c}$ into a lower-dimensional latent space $\boldsymbol{z}$, and the decoder generates new solutions $\boldsymbol{X}'$ conditioned on $\boldsymbol{c}$.

\par Once the CVAE is trained, it can generate initial candidate solutions for a new optimization instance. Specifically, given the current environmental factors $\boldsymbol{c}_{\text{new}}$, a latent variable $\boldsymbol{z}$ is sampled from the prior distribution $p(\boldsymbol{z})$, and the decoder generates a new candidate solution \(\boldsymbol{X}'\) by modeling \(p_{\theta}(\boldsymbol{X}' | \boldsymbol{z}, \boldsymbol{c}_{\text{new}})\). This approach allows the generation of high-quality initial solutions that are well-adapted to the current environmental conditions, thus improving the efficiency and effectiveness of the optimization process.

%
\subsection{GenSI Evolution}
\label{ssec:gensi_evolution}

\par After the initialization process gets the candidate solutions, the evolution phase of the GenSI framework seeks to optimize the candidate solutions to converge towards the optimal set of trade-offs. The evolution phase of the GenSI framework is based on the multi-objective ant lion optimizer (MOALO) algorithm, and thus we first introduce the evolution process of MOALO and then present the proposed enhancements to improve convergence and solution quality.

%
\subsubsection{MOALO Evolution Method}

\par MOALO is one of the state-of-the-art swarm intelligence algorithms, and as a branch of artificial intelligence, MOALO  can potentially solve some NP-hard problems~\cite{Tang2021}. The evolution process based on MOALO in GenSI involves the following steps:

\begin{enumerate}
  \item Calculate Objective Functions: The candidate solutions generated by the GenSI initialization process will be evaluated by calculating the objective values by using Eqs.~\eqref{eq:AF}-\eqref{eq:formulation}. Mathematically, we let $F_{n}=[f_{1,n}, f_{2,n}, f_{3,n}]$ denoting the three optimization objective values of the $n$th candidate solutions. 
  
  \item Compare and Reserve Solutions: MOALO maintains an archive $\mathcal{A}=\{\boldsymbol{X}_1^A, \boldsymbol{X}_2^A, ...\}$ to store elite solutions from previous iterations. During each iteration, the algorithm merges the current population with the archive via $\mathcal{A} \leftarrow \{\mathcal{A}, \mathcal{P}\}$. Solutions demonstrating superior objective values (\textit{i.e.}, non-dominated solutions defined below) are preserved in $\mathcal{A}$. Given the nature of MOPs, arithmetic relational operators cannot compare different solutions. Thus, we adopt Pareto dominance to prioritize solutions. 

  \begin{definition} [Pareto dominance]
  $\boldsymbol{X}$ \textit{Pareto dominance} $\boldsymbol{X}'$ iff: $\left[\forall o \in\{1,2,3\},f_o(\boldsymbol{X}) \leq f_o(\boldsymbol{X}')\right] \wedge \left[\exists o \in\{1,2,3\} f_o(\boldsymbol{X}) < f_o(\boldsymbol{X}')\right]$.
  \end{definition}

  \par The non-dominated solutions reserved by the archive can be defined as follows. 
  \begin{definition} [Non-dominated solution]
  $\boldsymbol{X}$ is called non-dominated solution iff: $\exists\mkern-13mu/ \boldsymbol{X'} \in \mathcal{A}$ Pareto dominance $\boldsymbol{X}$.
  \end{definition}

  \par  When the archive reaches capacity, a crowding mechanism removes solutions with similar trade-offs, ensuring diverse coverage of possible trade-offs.

  \item Update Candidate Solutions: MOALO selects candidate solutions from $\mathcal{A}$ by using the roulette wheel selection. Then, MOALO uses the selected candidate solutions to update the candidate solutions as follows:
  \begin{equation}\label{eq:solution-update}
  	\boldsymbol{X}_{n}=(\boldsymbol{X}^{R}+\boldsymbol{X}^{A})/2,
  \end{equation}
  \noindent where $\boldsymbol{X}^{R}$ is the guide solution calculated by heuristic principles of MOALO~\cite{Mirjalili2017b}, and $\boldsymbol{X}^{A}$ is the solution selected from $\mathcal{A}$.

  \item Terminate Algorithm: Determine whether the termination condition is reached. If not met, repeat steps 1) -4). Otherwise, return solutions in $\mathcal{A}$ as final results.
\end{enumerate}

\par MOALO encounters several challenges when utilized within the evolution process of the GenSI framework. \textit{Firstly}, in addressing our MOP, the non-dominated solutions generated by MOALO may not effectively serve our problem requirements, particularly when solutions exhibit extreme objective biases (\textit{e.g.}, optimizing one objective while significantly compromising the others). For example, a solution achieving minimal energy consumption but negligible security performance would be impractical. \textit{Secondly}, while our problem incorporates integer decision variables, MOALO is designed to handle only continuous variables. Therefore, we propose several enhancements to address these limitations in MOALO.

%
\subsubsection{Proposed Improvements to MOALO}
\label{sssec}

\par In this part, we propose two improvements to MOALO for boosting the evolution process. 

\par \textbf{1) Sorting-Based Population Evolution:} We first propose a novel scheme to improve the evolution of the population. We begin by proving a relationship between the archive and population evolution. 

\par Specifically, the distribution of solutions in the archive determines the search direction of the algorithm. In this case, MOALO may waste computational resources on excessively biased trade-offs, where one objective is prioritized while disregarding the other two objectives. The reason is that the solutions in the current population are derived from both the solutions in the archive and the solutions from the previous population in Step 3) of MOALO. Due to the crowding mechanism outlined in Step 2), MOALO tends to favor a diverse range of trade-offs between objectives. Consequently, the archive may contain solutions that exhibit highly biased trade-offs. These biased solutions can misguide the search process of MOALO in subsequent iterations.

\par Considering this property, we aim to eliminate less efficient trade-offs by filtering candidate solutions of the archive. Our sorting-based population evolution method is as follows. 

\par \textit{Firstly}, in each iteration, we rank the populations separately according to the three optimization objectives. \textit{Secondly}, we record the minimum value of the first and second optimization objectives and the maximum value of the third optimization objective, which are denoted as $f_{1_{min}}^{t}$, $f_{2_{min}}^{t}$, and $f_{3_{max}}^{t}$, respectively. \textit{Finally}, we remove the candidate solutions below the set three thresholds. The three thresholds $\zeta_1^t$, $\zeta_2^t$, and $\zeta_3^t$ are given by
\begin{equation}\label{eq:zeta}
	\begin{aligned}
		\zeta_1 = f_{1_{min}}^{t} \times {\delta_1}, \quad
		\zeta_2 = f_{2_{min}}^{t} \times {\delta_2}, \quad
		\zeta_3 = f_{3_{max}}^{t} \times {\delta_3},
	\end{aligned}
\end{equation}
\noindent where $\delta_1$, $\delta_2$, and $\delta_3$ are three weight parameters which are ranged from 0 to 1. If we require high coverage for a specific objective, we can set the corresponding parameters relatively larger. Following this, the main steps of the sorting-based population evolution method are shown in Algorithm~\ref{algo:Sorting}.

\begin{algorithm}[tb]
  \caption{Sorting-based Population Evolution}\label{algo:Sorting}
  \KwIn{$\mathcal{A}$, $\mathcal{P}$, $N$, current iteration $t$.}
  \KwOut{$\mathcal{A}$}
  $\mathcal{A} \leftarrow \{\mathcal{A}, \mathcal{P}\}$; (Denote the size of $\mathcal{A}$ as $N_A$ and the objectives of $a$th solution in $\mathcal{A}$ as [$f_{1,a}$,$f_{2,a}$,$f_{3,a}$])\\
  Rank $\mathcal{A}$ according to the first, second, and third objective values and record $f_{1_{min}}^{t}$, $f_{2_{min}}^{t}$, and $f_{3_{max}}^{t}$, respectively;\\

  Calculate $\zeta_1$, $\zeta_2$, and $\zeta_3$ by using Eq.~\eqref{eq:zeta};\\

  The dominated solutions are removed from $\mathcal{A}$;\\

  \For{$a=1$ to $N_A$}
  {
    \If {$f_{1,a}> \zeta_1$ and $mod(t,3)=0$ }
    {
    	The $a$th solution is removed from $\mathcal{A}$;\\
    }

    \If {$f_{2,a}> \zeta_2$ and $mod(t,3)=1$ }
    {
      The $a$th solution is removed from $\mathcal{A}$;\\
    }

    \If {$f_{3,a}> \zeta_3$ and $mod(t,3)=2$ }
    {
      The $a$th solution is removed from $\mathcal{A}$;\\
    }

  }
  Return $\mathcal{A}$;\\
\end{algorithm}

\par \textbf{2) Integer Update Method:} Another significant challenge of MOALO for solving our optimization problem lies in dealing with the integer decision variables. Specifically, the update method mentioned in Eq. \eqref{eq:solution-update} can update only the candidate solution with continuous decision variables. However, the formulated optimization problem involves integer decision variables. Thus, we propose an integer update method in the following. 

\par Akin to Eq.~\eqref{eq:solution-update}, we utilize the selected candidate solutions of the archive, \text{i.e.}, $\boldsymbol{X}^A$, to guide the update of MOALO. Moreover, we incorporate the integer decision variables of the original candidate solution $\boldsymbol{X}_n$ and a random integer to preserve inertia and increase randomness, respectively. Let $\boldsymbol{u}_{\mathcal{A}}= \boldsymbol{X}^A (\boldsymbol{u})$, $\boldsymbol{u}_{o}=\boldsymbol{X}_n (\boldsymbol{u})$, and $\boldsymbol{u}_{rand}=[randi(N_U), randi(N_U)]$ denote the integer decision variables of selected archive solution and original solution, and randomly generated integer, respectively, in which $randi(N)$ is a function that generates a random integer not exceeding $N$. Following this, the integer decision variables are updated as follows:
\begin{equation}
	\label{eq:integer-update}
	\boldsymbol{u}=\left\{\begin{array}{ll}
\boldsymbol{u}_{\mathcal{A}}, & rand<\frac{1}{3} \\
\boldsymbol{u}_{o}, & \frac{2}{3}>rand>\frac{1}{3} \\
\boldsymbol{u}_{rand}, & \text{otherwise} \\
\end{array}\right..
\end{equation}

\noindent As can be seen, the updated integer decision variables are guided by elite, inertia, and randomness mechanisms~\cite{Mirjalili2017b}, thereby achieving a more balanced search of the integer solution space.

\par By introducing the aforementioned sorting-based population evolution, and integer update methods, the main steps of the evolution process of GenSI are shown in Algorithm~\ref{algo:MOALO-RSI}, in which $N$ and $t_{max}$ are the population size and maximum iteration of the algorithm, respectively. Note that $t_{max}$ can be determined by historical experience. Additionally, in scenarios with limited computational resources, it is also possible to establish thresholds for the three objectives. This allows for early iteration termination when a solution exceeding these preset thresholds is discovered. Through the proposed improvements, the proposed GenSI can achieve more efficient evolution and better solution quality. As a result, GenSI can quickly explore a set of near-optimal solutions based on better initial solution quality.

%
\begin{algorithm}[b]
  \caption{Evolution Process of GenSI}\label{algo:MOALO-RSI}
  \KwIn{$\mathcal{P}$, $\mathcal{A}$, $N$, $t_{max}$ (maximum iteration), $\mathcal{P}_0$ (maximum iteration).}
  \KwOut{$\mathcal{A}$}

  $\mathcal{P} \leftarrow \mathcal{P}_0$, $\mathcal{A} \leftarrow \varnothing$;\\

  \For{$n=1$ to $N$}
  {
    Generate the solution $\boldsymbol{X}_{n}^{init}$ by using Eq. \eqref{eq:generate}; \\
    $\mathcal{P} \leftarrow \mathcal{P} \cup \left\{\boldsymbol{X}_{n}^{init}\right\}$;\\
  }
  \While{$t<t_{max}$}
  {
  	\For{$n=1$ to $N$}
	  {
	    Evaluate the objective values of $\boldsymbol{X}_{n}$ via Eqs.~\eqref{eq:AF}-\eqref{eq:formulation}, which are denoted as $F_{n}=[f_{1,n}, f_{2,n}, f_{3,n}]$;\\
	  }

	  Update $\mathcal{A}$ by using \textit{\textbf{Algorithm}} \ref{algo:Sorting};\\

  	  \For{$n=1$ to $N$}
	  {
	    Update $\boldsymbol{X}_{n}$ by using Eqs.~\eqref{eq:solution-update} and~\eqref{eq:integer-update};\\
	  }
    $t=t+1$;\\
  }
  Return $\mathcal{A}$;\\
\end{algorithm}

%
\subsection{Computational Complexity of GenSI}
\label{sssec:moalo_evolution}

\par The computational complexity of GenSI frameworks are analyzed as follows:

\begin{itemize}
    \item \textbf{GenSI Initialization:} The initialization process is divided into two scenarios, which are cold start and warm start.

    \par \textit{Cold Start:} In the cold start scenario, the random walk-based initialization is used to generate the initial candidate solutions. This method has a complexity of $\mathcal{O}(N)$, where $N$ is the population size, providing a diverse set of initial solutions.
    
    \par \textit{Warm Start:} In the warm start scenario, a CVAE is employed to generate the initial candidate solutions. The training complexity of the CVAE depends on the size of the dataset ($N_d$), the number of latent variables ($z$), and the depth of the neural network ($L$). Specifically, training the CVAE involves optimizing a variational lower bound for each data point, which requires backpropagation and iterative updates, resulting in a complexity of approximately $\mathcal{O}(N_d \cdot L \cdot z)$, where $N_d$ is the number of training samples, $L$ is the number of layers in the encoder and decoder networks, and $z$ is the dimension of the latent space. However, for generating the initial population, the complexity is $\mathcal{O}(N)$, as generating each sample involves a forward pass through the trained CVAE, which is significantly more efficient compared to the training phase.

    \item \textbf{GenSI Evolution:} The evolution phase is primarily based on the MOALO algorithm, which aims to iteratively refine the candidate solutions to identify the optimal trade-offs between multiple conflicting objectives. The computational complexity of MOALO can be analyzed as follows:

    \par The computations of the objective functions and the crowding mechanism determine the computational complexity of multi-objective optimization algorithms. If the number of optimization objectives is $N_o$, the computational complexity for calculating the objective functions is $\mathcal{O}(N_o N)$, where $N$ is the population size. The complexity for the crowding mechanism and non-dominated sorting is $\mathcal{O}(N_o N_{Arc} \log N_{Arc})$, where $N_{Arc}$ represents the size of the Pareto archive. In most cases, the size of the Pareto archive is the same as the population size $N$, which implies that the computational complexity for non-dominated sorting becomes $\mathcal{O}(N_o N^2)$. Therefore, the overall complexity for MOALO is $\mathcal{O}(N_o N^2)$.
    
    \par The improvements proposed in GenSI, such as sorting-based population evolution and the integer update method, do not increase the asymptotic complexity, keeping the evolution phase complexity at $\mathcal{O}(N_o N^2)$.

\end{itemize}

\par As such, the overall computational complexity of GenSI can be summarized as follows. The initialization phase contributes a smaller complexity of $\mathcal{O}(N)$, while the evolution phase dominates with a complexity of $\mathcal{O}(N_o N^2)$. Thus, the final time complexity of GenSI is $\mathcal{O}(N_o N^2)$.

%
\subsection{Deployment Scheme and Applicable Scenarios}
\label{sssec:deployment_scheme}

\par When the GenSI framework is deployed in practice, the GenSI framework necessitates a scheme for acquiring prior information and sharing results. We discuss a viable deployment scheme as follows:

\begin{itemize}

\item \textit{Acquiring Prior Information}: AAV swarms use non-optimized robust DCB to share crucial details, such as their positions, locations of identified eavesdroppers, and other prior information. Note that the non-optimized DCB enables long-range transmission but may not have sufficient secure measures, hence the encryption is used to ensure confidentiality.

\item \textit{Execution of GenSI}: The next step is running GenSI to obtain the optimization results. GenSI can be executed on a centralized high-performance computing device (such as a mobile terrestrial workstation) or deployed as a parallel distributed version that runs across computation devices within the AAV swarm. In a cold start scenario, more iterations are needed to achieve convergence due to the lack of prior knowledge, whereas in a warm start scenario, fewer iterations are required due to the high-quality initial solutions provided by the trained CVAE.

\item \textit{Sharing Optimization Results and Implementation}: The AAV swarm running GenSI sends the optimization results and schedules to other AAV swarms by using non-optimized robust DCB. Subsequently, the AAV swarms can proceed with DCB-based secure communication according to the optimization results generated by GenSI.

\end{itemize}

\par This deployment approach ensures a balance between practicality and computational efficiency. By leveraging distributed processing within the AAV swarm or centralized computing when accessible, the GenSI framework can efficiently adapt to different scenarios and resource constraints, ensuring secure and optimized AAV communication.

\par Note that the proposed method conducts optimization under relatively idealized conditions.  However, considering the practical effect, there are several limitations in practical applications, which are summarized as follows.

\begin{itemize}

\item \textit{Firstly}, the number of AAVs in each swarm should not be excessively large. This is since that an increased number of AAVs can lead to significant difficulty in practical controlling and synchronizing. 

\item \textit{Secondly}, our DCB-based method is more applicable when the distance between the AAV swarms is sufficiently large. This condition aligns with the advantages of the DCB method, which is designed to facilitate communication over longer distances. In short-distance scenarios, dispatching one AAV fly to another AAV swarm for communication by using the high maneuverability of the AAVs would be more effective

\item \textit{Finally}, the spacing between AAVs within each swarm should be compact. If the internal spacing is too large, it can increase the difficulty associated with maintaining synchronization.

\end{itemize}

%
%
\section{Simulation Results} 
\label{sec:simulation_results_and_analysis}

\par In this section, we provide extensive simulation results to evaluate the effectiveness of the proposed method.

\begin{table}
	\centering
	\caption{Numeral results obtained by baselines and our GenSI framework.}
	\label{tab:result-plos-small}
	\begin{tabular}{lllll}
		\toprule
		\textbf{Method} 	& \bf{$f_{1}$ [bps] }          & \bf{$f_{2}$ [dB] } & \bf{$f_{3}$ [J] } \\ \midrule
    LAA-Swarm               & $7.59 \times 10^{5}$    & $-0.21$          & $8.23 \times 10^{4}$  \\
    MOGOA~\cite{Mirjalili2018}                   & $1.90 \times 10^{6}$    & $-0.92$          & $1.04 \times 10^{5}$    \\
    MOMVO~\cite{Mirjalili2017}                    & $2.00 \times 10^{6}$    & $-1.73$          & $1.11 \times 10^{5}$   \\
    MSSA~\cite{Mirjalili2017a}                     & $1.93 \times 10^{6}$    & $-0.48$          & $8.08 \times 10^{4}$   \\
    MODA~\cite{Mirjalili2016}                     & $1.98 \times 10^{6}$    & $-2.32$          & $9.41 \times 10^{4}$  \\
    MOALO~\cite{Mirjalili2017b}       & $2.01 \times 10^{6}$    & $-1.63$          & $1.04 \times 10^{5}$   \\
    Our GenSI (Cold Strat)               & \bm{$2.14 \times 10^{6}$}    & \bm{$-2.48$}      & \bm{$7.14 \times 10^{4}$} \\
    
    Our GenSI (Warm Strat)               & \bm{$2.12 \times 10^{6}$}    & $-2.11$      & $7.68 \times 10^{4}$ \\
    \bottomrule
	\end{tabular}
\end{table}

%
\begin{figure*}
    \centering
    \subfloat[]{
       \includegraphics[width=0.245\linewidth]{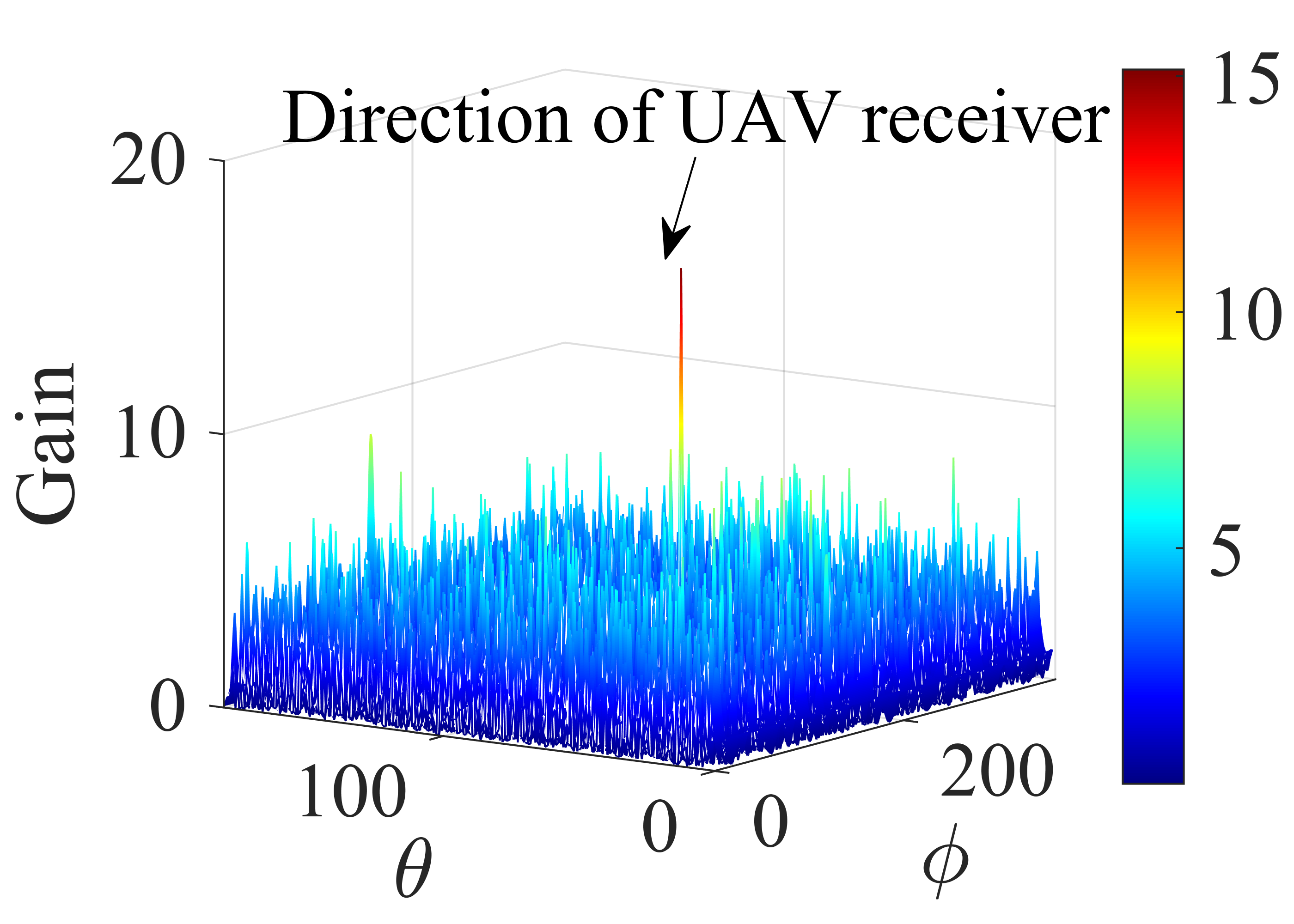}\label{fig:vr-gain1}}
    \subfloat[]{
       \includegraphics[width=0.245\linewidth]{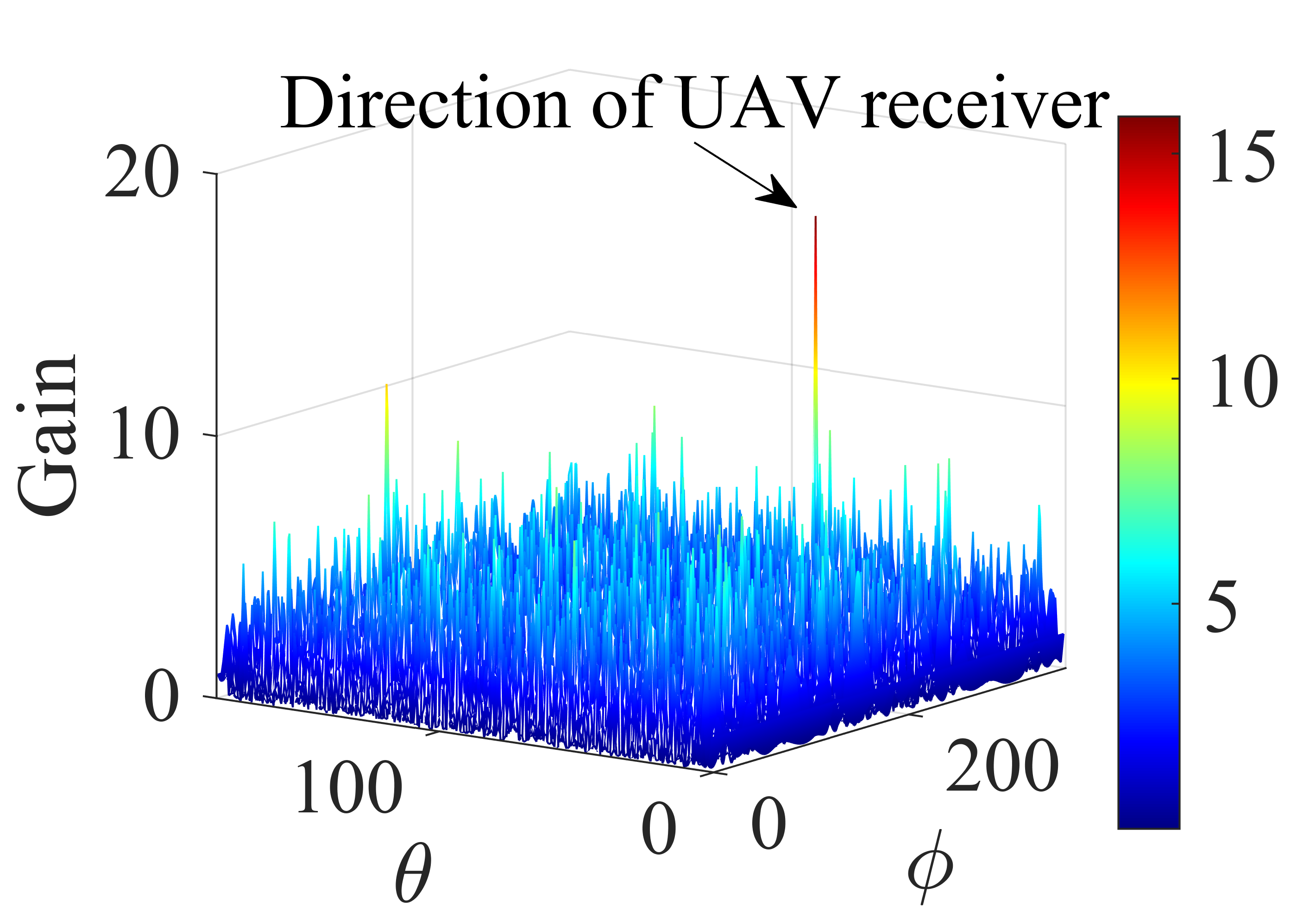}\label{fig:vr-gain2}}
        \subfloat[]{
       \includegraphics[width=0.245\linewidth]{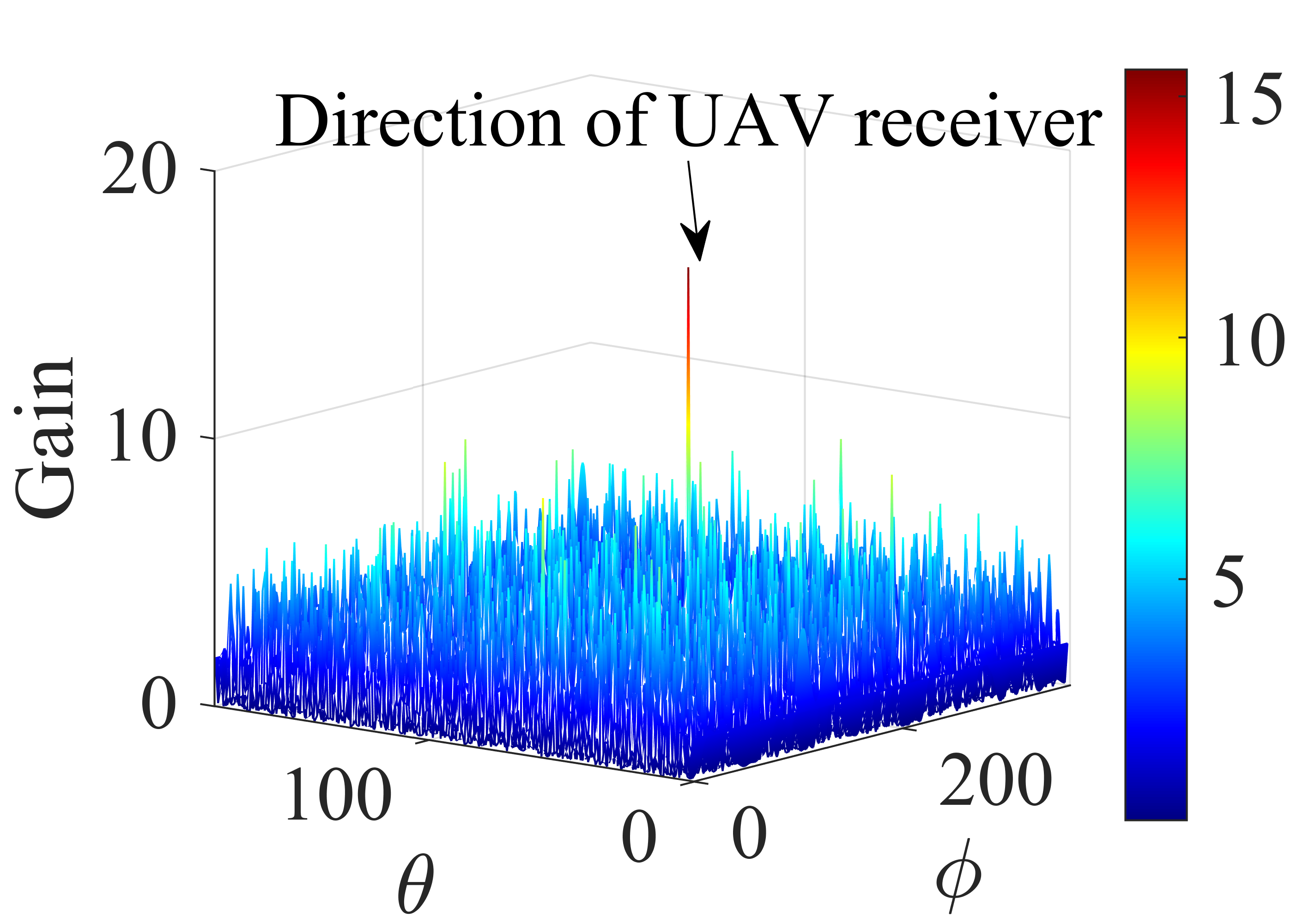}\label{fig:vr-gain3}}
    \subfloat[]{
       \includegraphics[width=0.245\linewidth]{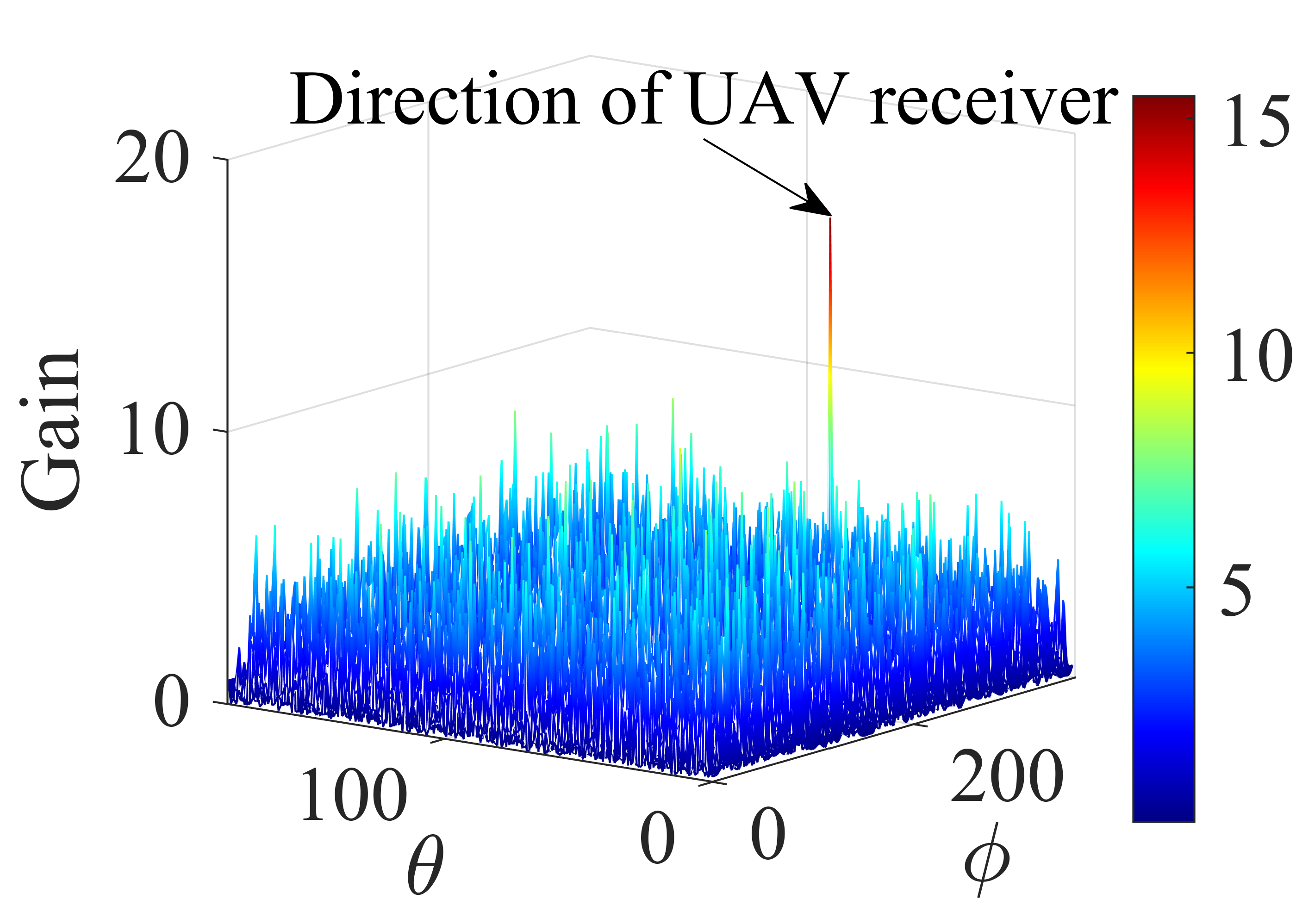}\label{fig:vr-gain4}}
  \caption{Antenna gains obtained by our proposed GenSI framework. (a) Antenna gain of AVAA1 in cold start case. (b) Antenna gain of AVAA2 in cold start case. (c) Antenna gain of AVAA1 in warm start case. (d) Antenna gain of AVAA2 in warm start case. }
  \label{fig:Visualization-results}
\end{figure*}

%
\begin{figure*}
    \centering
    \subfloat[Cold start case]{
    \includegraphics[width=0.485\linewidth]{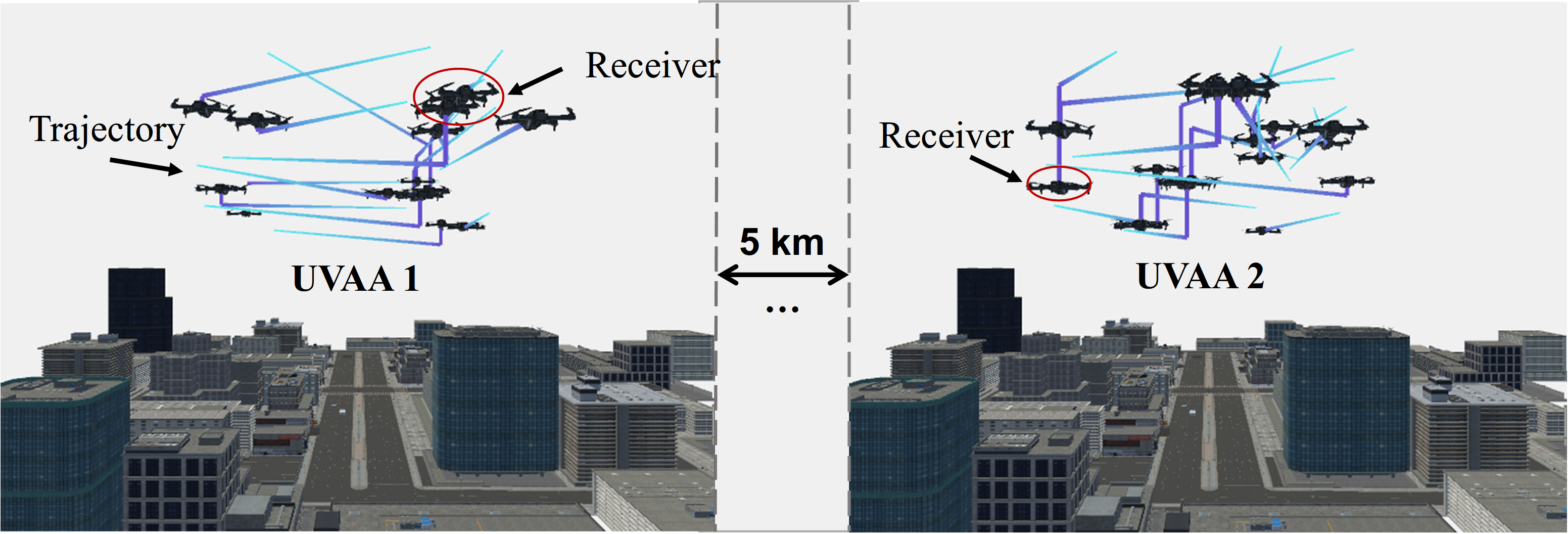}\label{fig:vr-path1}}
    \subfloat[Warm start case]{
    \includegraphics[width=0.485\linewidth]{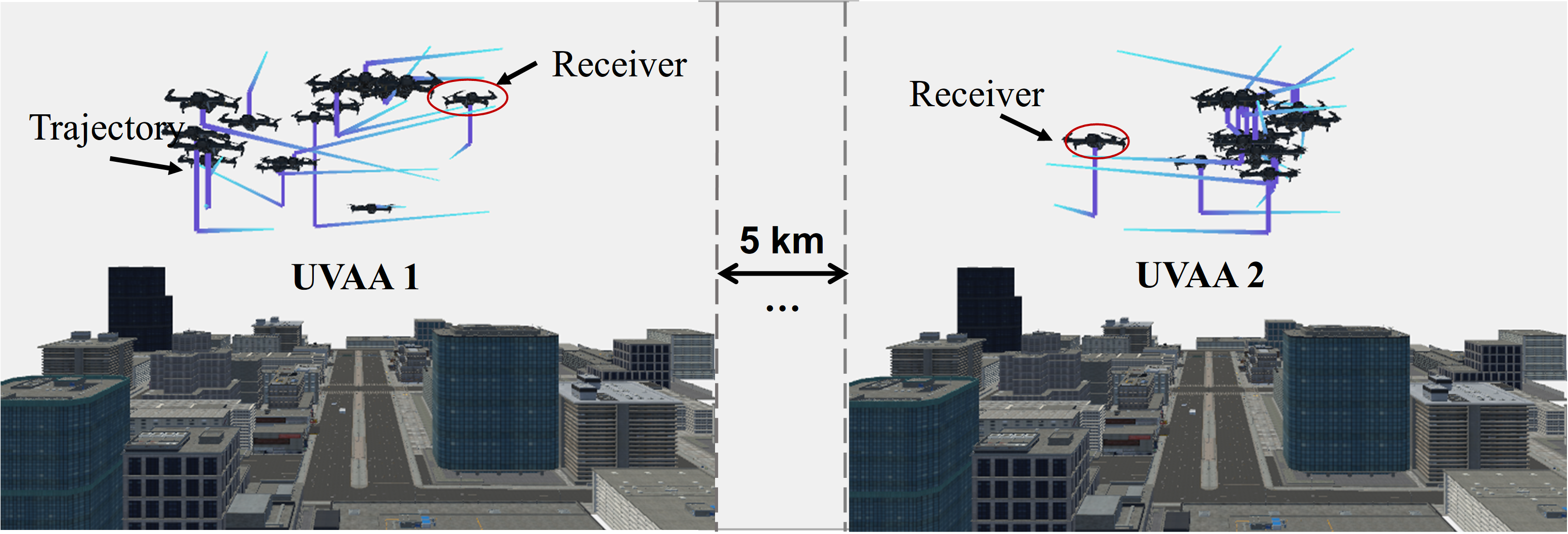}\label{fig:vr-path2}}
   
  \caption{Trajectories of the AAV swarms obtained by our proposed GenSI framework.}
  \label{fig:path-results}
\end{figure*}

%
\subsection{Simulation Setups} 

\par In the simulations, the numbers of AAVs of two AAV swarms are set as 16, and the AAV heights vary from 70 m to 120 m~\cite{Li2024}. Moreover, these two AAV swarms are separately distributed in two 100 m $\times$ 100 m areas, and the distance between these two areas is about 5 km~\cite{Li2023}. The collision distance $d_{min}$ between two arbitrary AAVs is set as 0.5 m~\cite{Li2024}. In addition, we select 0.9 GHz as the operating frequency because it offers a balanced combination of long-range communication and bandwidth with low interference, making it suitable for our scenario, and existing practical DCB deployments have demonstrated its feasibility at this frequency~\cite{Mohanti2022,Mohanti2019} (The details are shown in Section~\ref{sec:frequency}). Other parameters related to communications and the AAV energy model follow~\cite{Mozaffari2019} and~\cite{Li2024}, respectively. For comparison, we adopt the following baselines:  
\begin{itemize}
    \item \textit{LAA-Swarm:} Two AAV swarms separately form two linear antenna arrays and randomly select a receiver from a different AAV swarm. 

    \item \textit{State-of-the-art Baseline Algorithms:} We consider state-of-the-art multi-objective swarm intelligence algorithms, including multi-objective grasshopper optimization algorithm (MOGOA)~\cite{Mirjalili2018}, multi-objective multi-verse optimizer (MOMVO)~\cite{Mirjalili2017}, multi-objective salp swarm algorithm (MSSA)~\cite{Mirjalili2017a}, multi-objective dragonfly algorithm (MODA)~\cite{Mirjalili2016}, and MOALO~\cite{Mirjalili2017b}. Note that these algorithms employ the proposed integer update method to handle the integer decision variables of the formulated problem. 
\end{itemize}

\par Moreover, all population-based swarm intelligence algorithms are configured with the population size and maximum iteration at 50 and 500, respectively, while their key parameters follow the initial settings of their source papers. As for the proposed GenSI framework, the maximum iteration is set as 500 in the cold start scenario while set as 200 in the warm start scenario. Note that the CVAE model in the warm start part of GenSI is trained on a dataset derived from 300 problem instances. This CVAE model employs the linear neural network architecture and a latent dimension of 20 to capture the underlying patterns between environmental factors and the optimal solutions.

%
\begin{figure}
  \centering
  \includegraphics[width=3.5 in]{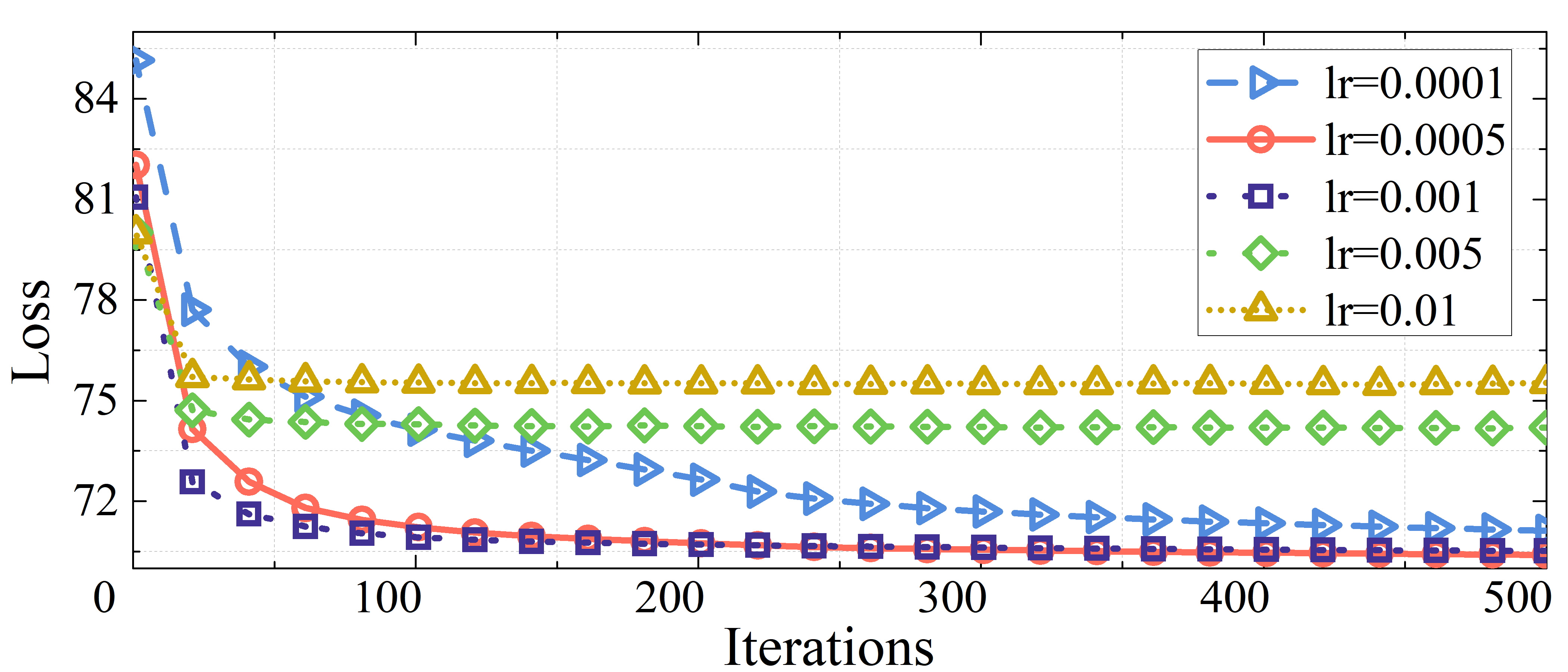}
  \caption{Training loss curves of GenSI framework under different learning rates, where learning rate (\textit{i.e.}, lr) of 0.0005 achieves the optimal convergence performance.}
  \label{fig:Loss}
\end{figure}

%
\begin{figure}
  \centering
  \includegraphics[width=3.5 in]{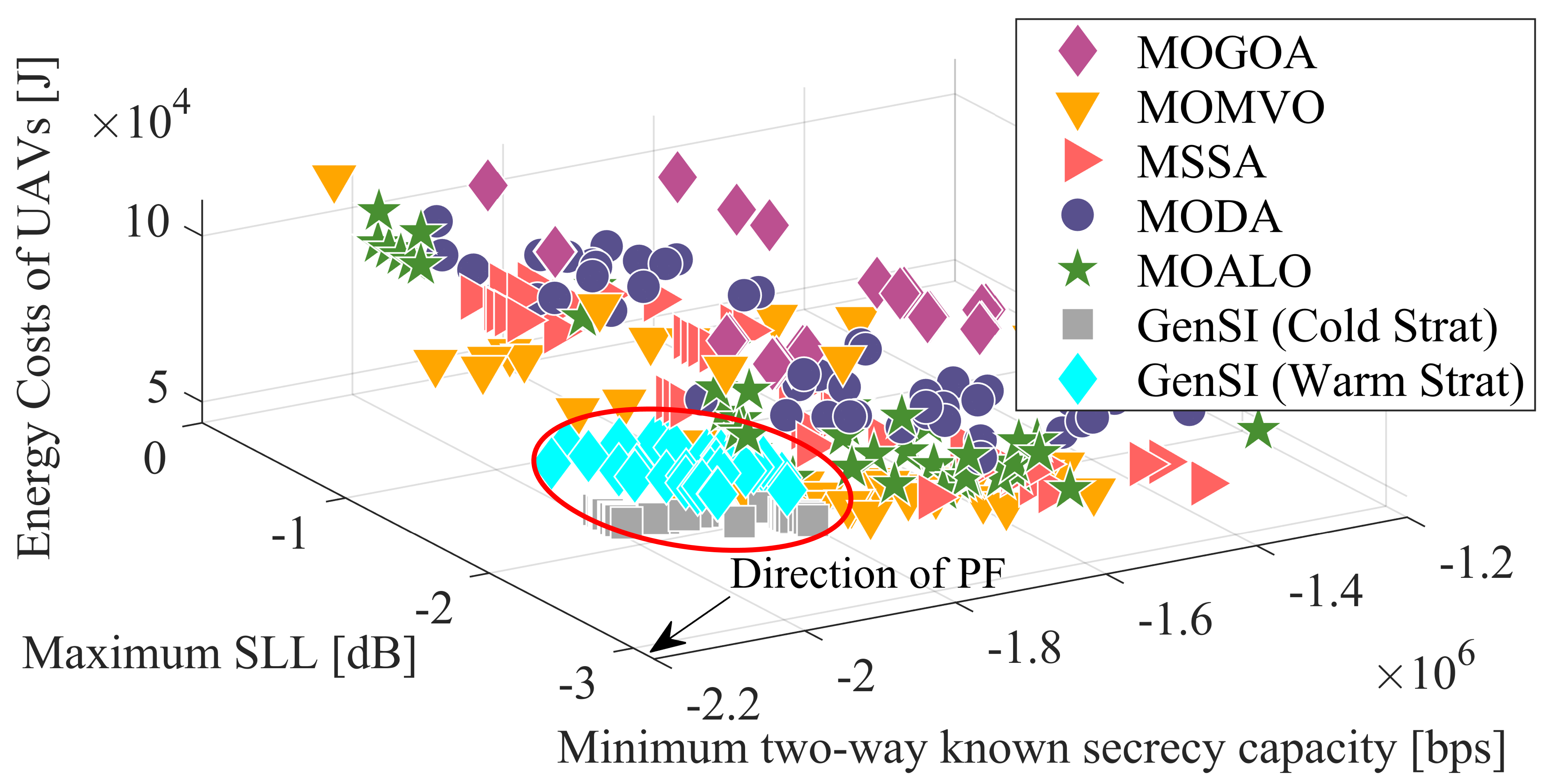}
  \caption{Pareto solutions obtained by benchmarks and the proposed GenSI framework. All the nodes denote the objective values of the candidate solutions obtained by different algorithms.}
  \label{fig:PS}
\end{figure}

%
\begin{figure*}
    \centering
    \subfloat[Phase synchronization error \\ (Cold start)]{
       \includegraphics[width=0.22\linewidth]{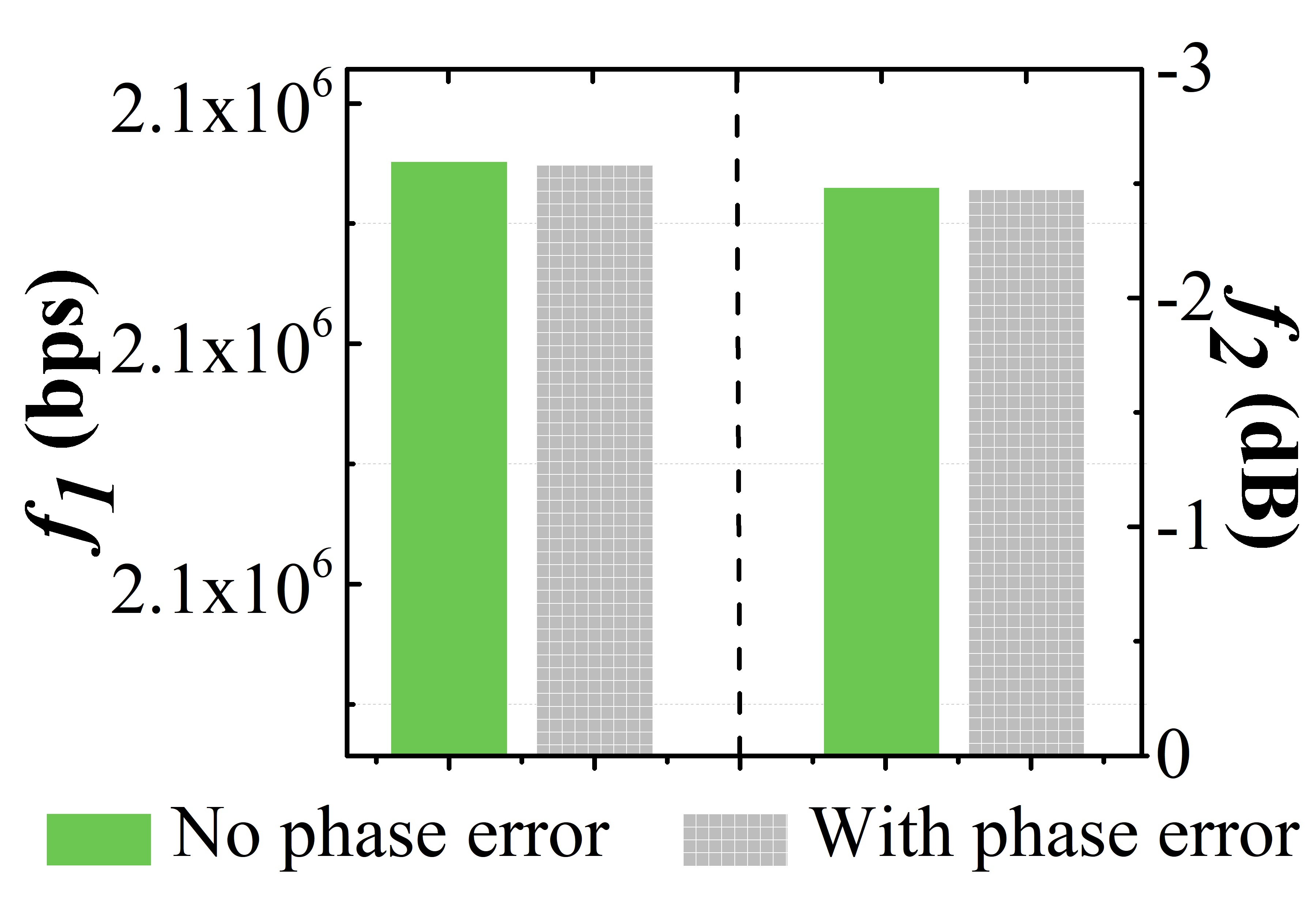}\label{fig:rv-phase1}}
    \subfloat[CSI errors (Cold start)]{
       \includegraphics[width=0.37\linewidth]{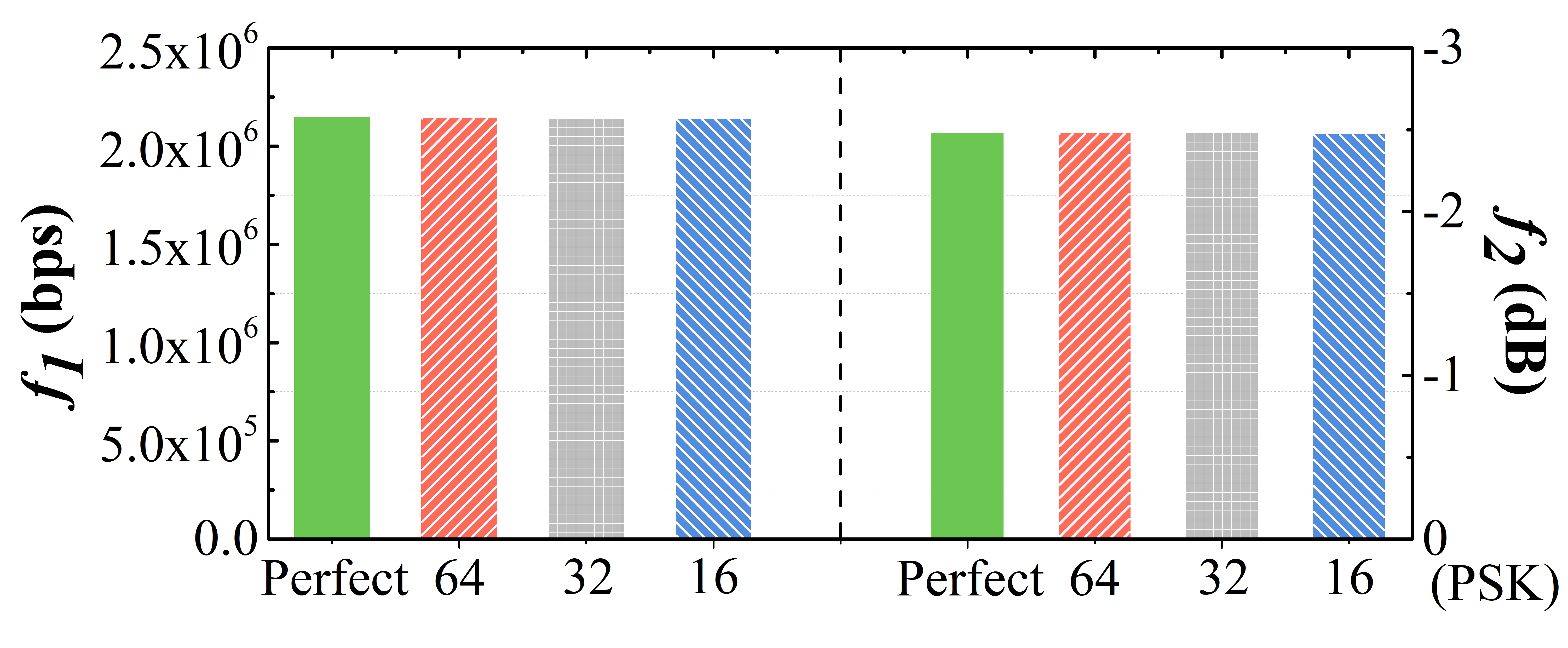}\label{fig:rv-CSI1}} 
     \subfloat[AAV jitter (Cold start)]{
       \includegraphics[width=0.37\linewidth]{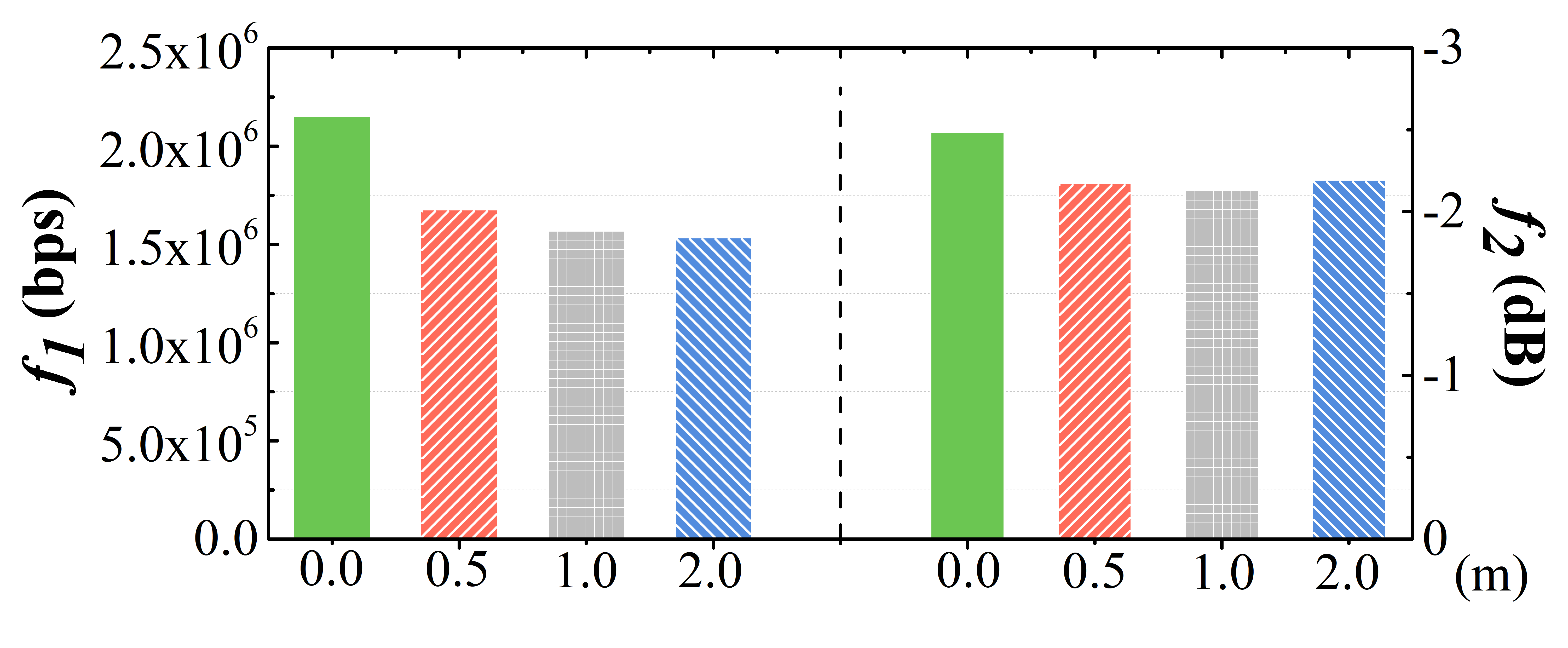}\label{fig:rv-position1}}
       \\

    \subfloat[Phase synchronization error (Warm start)]{
       \includegraphics[width=0.22\linewidth]{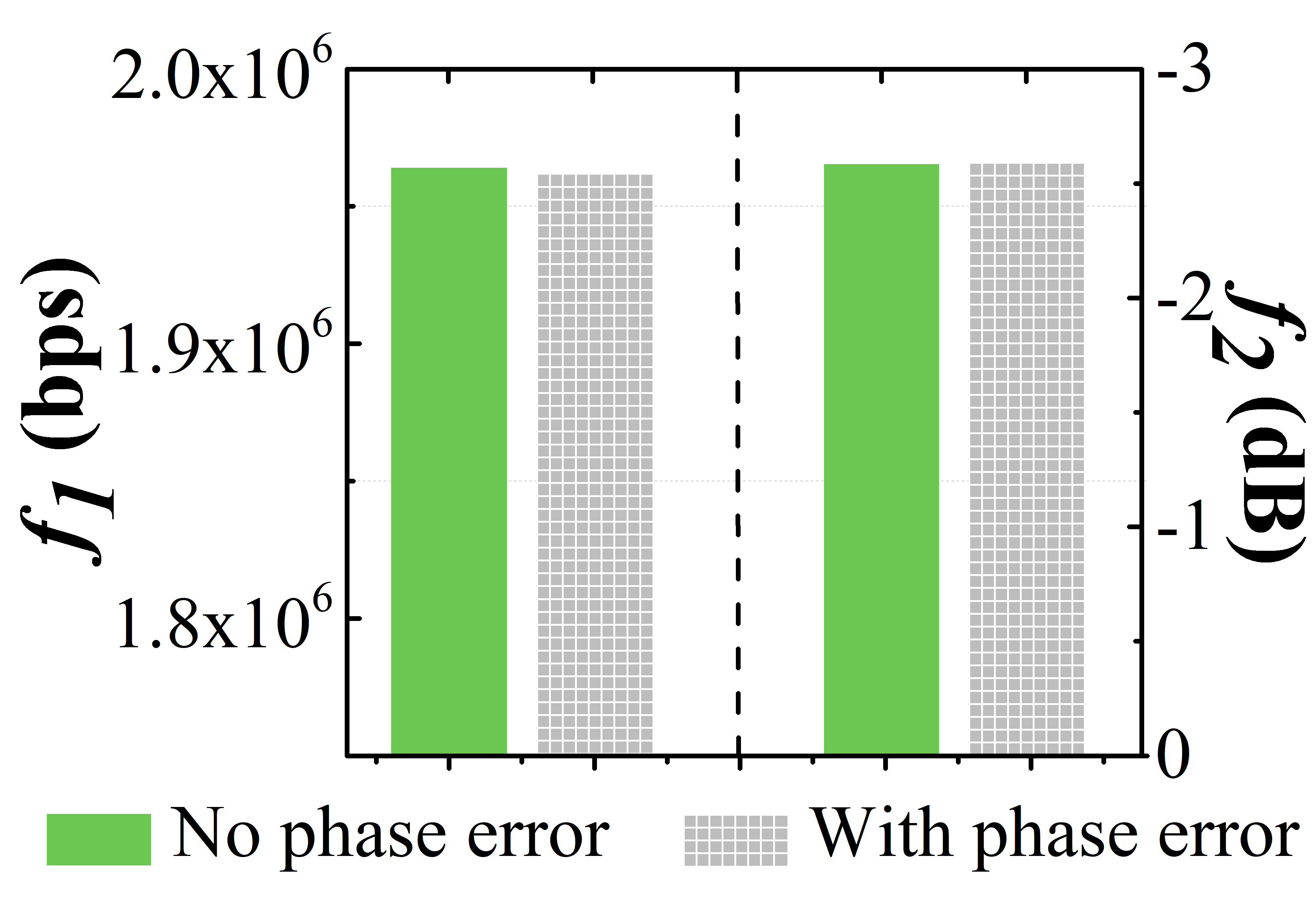}\label{fig:rv-phase2}}
    \subfloat[CSI errors (Warm start)]{
       \includegraphics[width=0.37\linewidth]{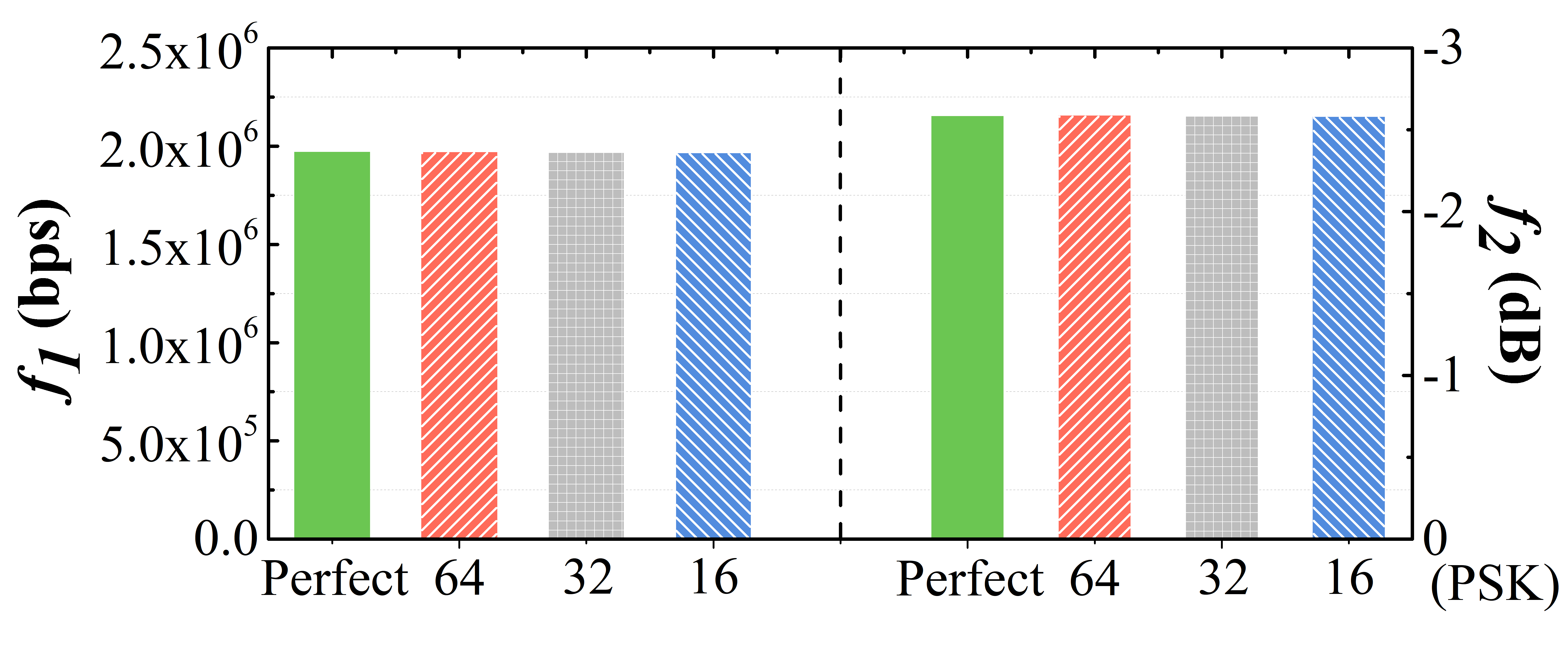}\label{fig:rv-CSI2}} 
     \subfloat[AAV jitter (Warm start)]{
       \includegraphics[width=0.37\linewidth]{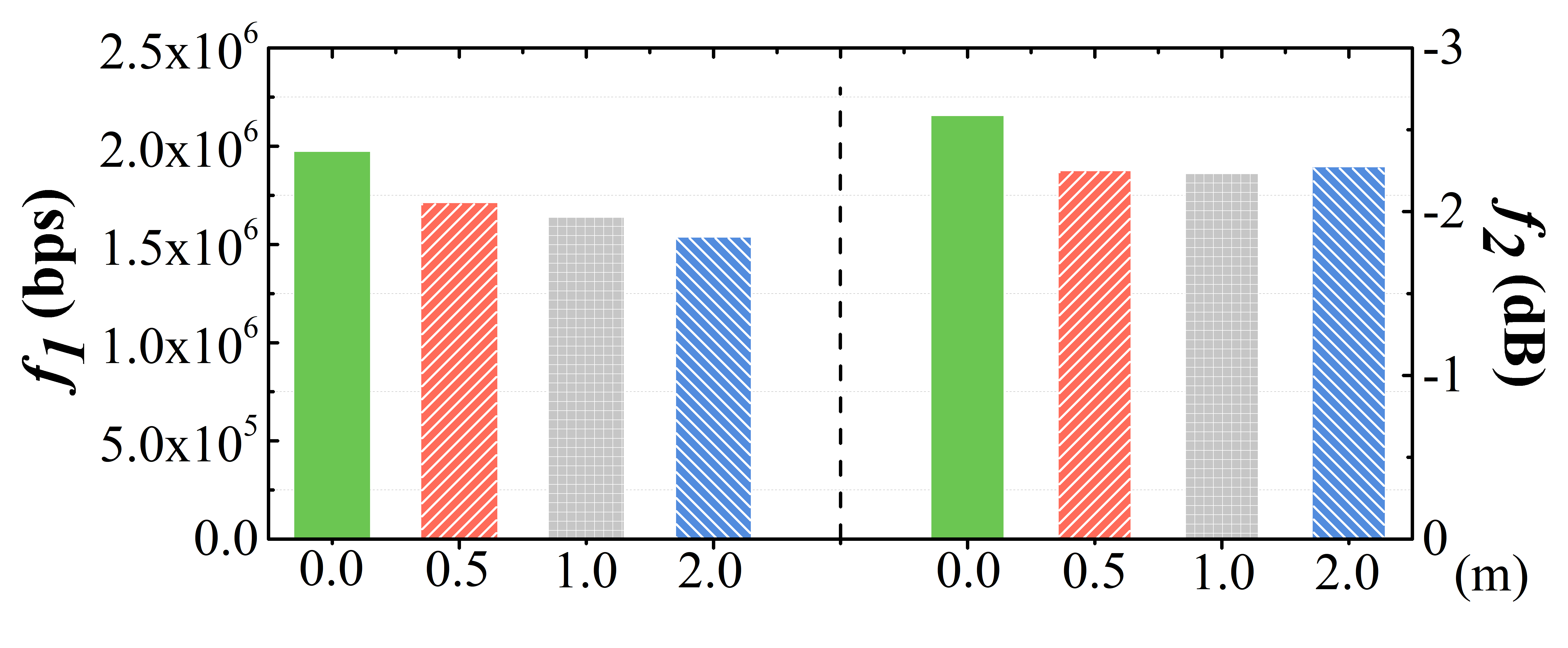}\label{fig:rv-position2}}
       \\
  \caption{Robustness verification results (in terms of $f_1$ and $f_2$) under some special cases.}
  \label{fig:Robustness-results}
\end{figure*}

%
\subsection{Convergence and Visualization Results} 

\par Fig.~\ref{fig:Loss} shows the training loss curves of the CVAE model of the GenSI framework under different learning rates, where learning rate (\textit{i.e.}, lr) are set as 0.01, 0.005, 0.001, 0.0005, and 0.0001, respectively. As can be seen, lr = 0.0005 achieves the optimal convergence performance, demonstrating both faster convergence rate and better stability compared to other learning rates. Based on the convergence results, we adopt lr = 0.0005 as the optimal learning rate for all subsequent experiments in this work to ensure both training efficiency and model stability.

\par Moreover, for ease of presentation, we employ Unity 3D and Matlab to visualize the results and demonstrate the effectiveness of the solution obtained by our GenSI framework. Firstly, we present the antenna gains of the AVAAs obtained by both cold start and warm start cases in Figs.~\ref{fig:Visualization-results}\subref{fig:vr-gain1},~\ref{fig:Visualization-results}\subref{fig:vr-gain2},~~\ref{fig:Visualization-results}\subref{fig:vr-gain3}, and~\ref{fig:Visualization-results}\subref{fig:vr-gain4}, respectively. It is evident that, except for the target directions, the antenna gains of both AVAAs are relatively low. This indicates that two-way aerial communications can achieve high secrecy capacities. Secondly, in Figs.~\ref{fig:path-results}\subref{fig:vr-path1} and~\ref{fig:path-results}\subref{fig:vr-path2}, we illustrate the trajectories of the AAVs in the two AAV swarms during constructing AVAAs. Notably, the AAVs exhibit minimal position changes, resulting in energy savings. Thus, these visualization results can be evidence that our solution achieves relatively high energy efficiency.


%
\vspace{+0.5mm}
\subsection{Comparisons and Analyses}

\par In this part, the GenSI framework is compared with other baseline methods mentioned in simulation setups in solving our optimization problem. Table \ref{tab:result-plos-small} shows the numeral results from our GenSI framework and the comparison benchmarks in terms of the three objectives as given in Eqs.~\eqref{eq: objecitve1},~\eqref{eq: objecitve2} and~\eqref{eq: objecitve3} (\textit{i.e.}, $f_1$, $f_2$, and $f_3$). For ease of analysis, we assume $f_1$ is the most crucial objective of this work, and select the solution with the best $f_1$ value from the Pareto solution sets as the final solution by using the automatic method in~\cite{Ferreira2007}. As can be seen, both the cold start and warm start cases of the GenSI framework are superior to the LAA-Swarm method which is most likely to be employed in practice, indicating that the considered optimization approach is non-trivial. Moreover, both the cold start and warm start cases of the GenSI framework outmatch various multi-objective swarm intelligence algorithms, implying that it is more suitable for solving the formulated problem. Thus, the formulation and the proposed enhanced measures of the GenSI framework are valid and effective. 

\par Fig. \ref{fig:PS} illustrates the Pareto solutions (\textit{i.e.}, candidate solutions in archive) generated by our GenSI framework alongside other baseline algorithms. As can be seen, the candidate solutions obtained by the GenSI framework are approaching the ideal Pareto front direction, and mostly dominate the candidate solutions obtained by other baseline algorithms. Furthermore, GenSI successfully avoids extreme trade-off biases (\textit{e.g.}, the trade-off with -0.02 dB SLL obtained by MOALO). This balanced distribution is achieved through our sorting-based population evolution approach, which systematically eliminates excessively biased trade-offs during each iteration.

\par Notably, we observe that the warm start case of GenSI achieves comparable optimization performance to the cold start case while requiring less than half of the iterations. This significant reduction in computational overhead while maintaining performance demonstrates the effectiveness of leveraging generative AI to learn and exploit environmental parameter patterns. Such findings suggest that the proposed warm start mechanism successfully captures and utilizes the underlying structure of the optimization problem, enabling more efficient convergence without compromising solution quality.

%
\subsection{Robustness Verification}

\par In this part, we examine the robustness of the proposed method in various practical unexpected circumstances. \textit{Firstly}, we model the phase synchronization error according to~\cite{Ahmad2022}, which follows a Gaussian distribution with zero-mean and variance $\zeta^{2}=\omega_{c}^{2} q_{1}^{2} \Delta T+\omega_{c}^{2} q_{2}^{2} \Delta T^{3}/3$ (Note that the parameters follow~\cite{Ahmad2022}). Figs.~\ref{fig:Robustness-results}\subref{fig:rv-phase1} and~\ref{fig:Robustness-results}\subref{fig:rv-phase2} show that our solution exhibits negligible performance degradation with such phase synchronization errors. \textit{Secondly}, we assess the impact of CSI errors that are induced when using different length CSI codebooks from~\cite{Ahmad2022}, including errors of 16-PSK, 32-PSK, and 64-PSK codebooks. Note that a longer codebook tends to yield smaller CSI errors. As depicted in Figs.~\ref{fig:Robustness-results}\subref{fig:rv-CSI1} and~\ref{fig:Robustness-results}\subref{fig:rv-CSI2}, the performance degradation in terms of $f_1$ and $f_2$ is generally insignificant in most cases, particularly when using the codebooks longer than 32-PSK. \textit{Finally}, we examine four AAV jitter conditions, in which the maximum drifts are set to 0.5 m, 1 m, and 2 m, respectively. As can be seen from Figs.~\ref{fig:Robustness-results}\subref{fig:rv-position1} and~\ref{fig:Robustness-results}\subref{fig:rv-position2}, the performance losses between the non-drift and position-drifted cases are relatively minimal for $f_2$. However, for $f_1$, there is a moderate degradation when drifts are present, although it remains acceptable for relatively small drifts. As such, the proposed method demonstrates a certain degree of robustness to overcome the potential unexpected circumstances.

%
\subsection{Impacts and Determinations of Carrier Frequency}
\label{sec:frequency}

\begin{figure}
\centering
\includegraphics[width=0.9\linewidth]{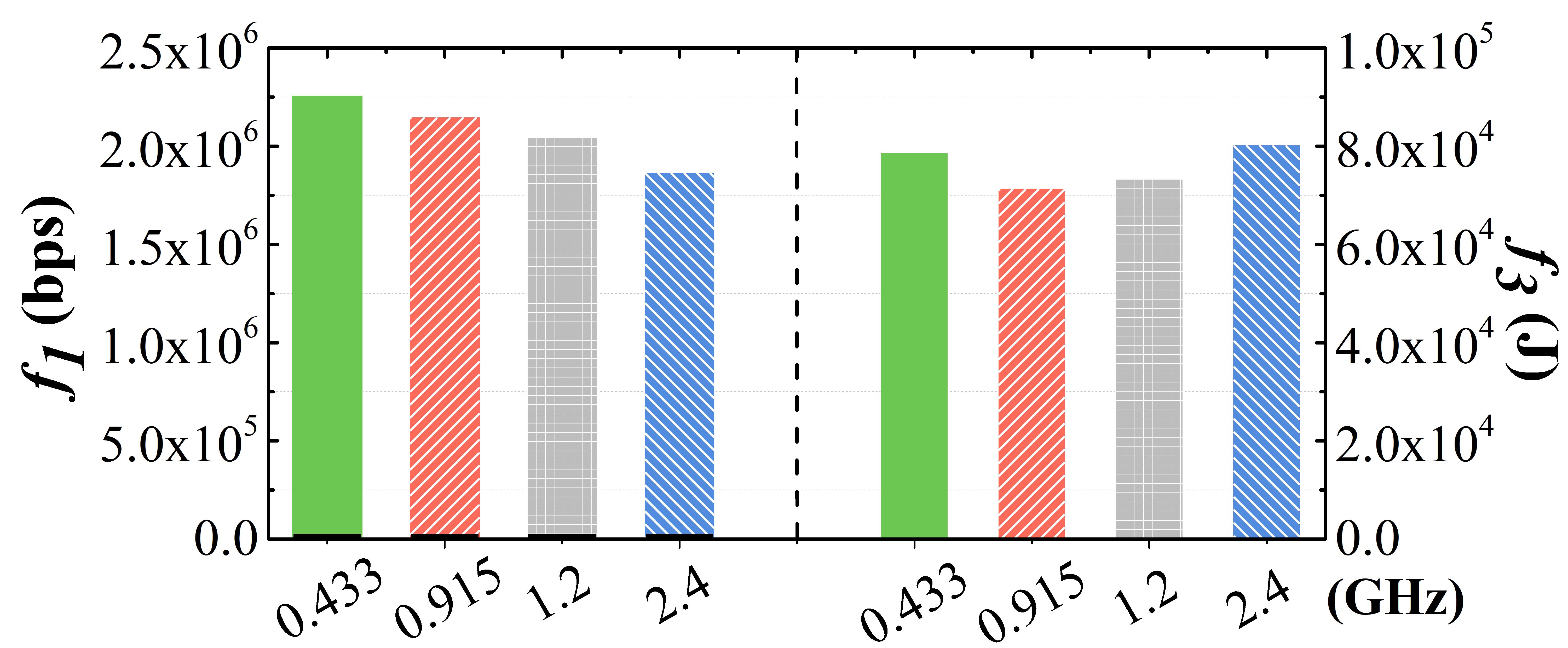}
\caption{Performance comparison of different operating frequencies (433 MHz, 915 MHz, 1.2 GHz, and 2.4 GHz).}
\label{fig:frequency}
\end{figure}

\par The operating frequency significantly influences the performance of AAV communication systems. To comprehensively evaluate this impact, we analyze four commonly used frequencies, which are 433 MHz, 915 MHz, 1.2 GHz, and 2.4 GHz. 

\par Note that each of these frequencies exhibits distinct characteristics that make them suitable for different AAV applications. Specifically, the 433 MHz frequency features superior penetration and long-range capabilities but limited bandwidth, making it ideal for agricultural monitoring and industrial inspection. Moreover, the 915 MHz frequency provides a balanced trade-off between range and bandwidth, with favorable penetration characteristics and relatively low interference, suitable for long-range operations such as delivery and power line inspection. In addition, the 1.2 GHz frequency offers increased bandwidth but experiences greater signal attenuation, commonly used in first-person view (FPV) racing and professional cinematography. Finally, the 2.4 GHz frequency, while providing high bandwidth, suffers from significant signal attenuation over distance, primarily utilized in urban photography and short-range applications.

\par To evaluate the performance of these frequency bands, we conduct simulations under equivalent bandwidth conditions. Fig.~\ref{fig:frequency} presents the simulation results, demonstrating that lower frequencies generally achieve better performance in long-distance transmission scenarios. This can be attributed to the longer wavelengths of lower frequencies, which are more conducive to long-range propagation.

%
\begin{figure}
  \centering
  \includegraphics[width=3.5 in]{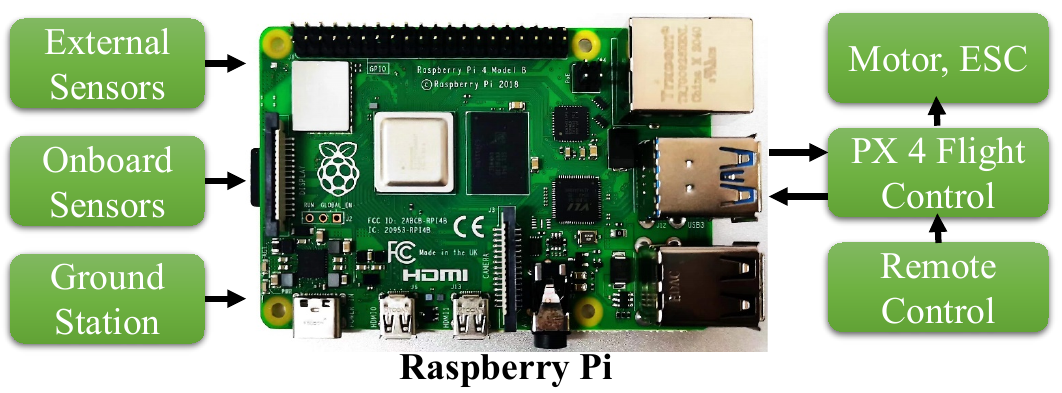}
  \caption{A calculation module of autonomous AAV systems based on Raspberry Pi 4B.}
  \label{fig:flight_control}
\end{figure}

\par Based on our analysis and considering the target scenario of long-range communication, we select and recommend 915 MHz as the operating frequency for our system. This carrier frequency provides adequate bandwidth and experiences less interference. Furthermore, experimental deployments have validated the effectiveness of DCB in AAV networks at this frequency~\cite{Mohanti2022,Mohanti2019}.

%
\subsection{Practicality Analysis} 

\par We evaluate the practicality of the proposed method against traditional encryption/decryption approaches. As illustrated in Fig.~\ref{fig:flight_control}, we employ the Raspberry Pi 4B as the AAV flight control system, since Raspberry Pi 4B is a common configuration in standard AAV platforms (\textit{e.g.}, PX4 autopilot) and previous studies (\textit{e.g.},~\cite{Zhou2021}). As mentioned in Section \ref{sec:algorithm}, we consider that one AAV swarm executes a parallel distributed version of the proposed GenSI framework, and we omit the step for calculating optimization objective values since it is usually substituted with proxy models~\cite{Jeong2005} in practical deployments. Moreover, three common encryption/decryption methods, namely, data encryption standard (DES), advanced encryption standard (AES), and Rivest-Shamir-Adleman (RSA), are introduced for the comparisons~\cite{Bhanot2015}.

\par Experimental results demonstrate that a one-time calculation of our GenSI framework can be completed within 65.26 s (Cold start case) or 26.98 s (Warm start case), respectively. In addition, the calculation times for encrypting and decrypting 200 MB of data by using DES, AES, and RSA are 12.07 s, 9.29 s, and 1567.59 s, respectively. Clearly, when the data volume exceeds approximately 1 GB, our proposed method with a cold start case achieves obvious advantages in computing time. This is because the encryption/decryption techniques need to continuously process over time while our method only needs one-time calculation. 

\par Moreover, the warm start case further enhances the practical applicability of physical layer security methods by reducing the computational overhead by approximately 58.7\% compared to the cold start case. This significant reduction in execution time makes our approach particularly attractive for real-time applications and resource-constrained scenarios. Moreover, the reduced computational burden enables more frequent security parameter updates, potentially improving the system resilience against emerging threats while maintaining minimal impact on overall system performance.


%
%
\section{Conclusion} 
\label{sec:conclusion}

\par This paper has investigated a DCB-enabled aerial two-way communication system between two AAV swarms under eavesdropper collusion in LAE scenarios. To achieve secure and energy-efficient communications, we have formulated a multi-objective optimization problem that simultaneously minimizes the two-way known secrecy capacity, the maximum sidelobe level, and the energy consumption of AAVs. The optimization aims to prevent information leakage to both known and unknown eavesdroppers while maintaining energy efficiency in constructing virtual antenna arrays. Due to the NP-hardness of the problem and the large number of decision variables, we have proposed a novel GenSI framework that combines an enhanced swarm intelligence algorithm with a CVAE-based generative method. The framework leverages the CVAE to learn from expert solutions in various environmental states and generate high-quality initial solutions for new scenarios. Simulation results have demonstrated that our proposed GenSI framework significantly outperforms state-of-the-art baseline algorithms. Furthermore, the integration of the CVAE mechanism substantially reduces computational overhead, thus decreasing execution time. Based on the existing far-field DCB framework, future research will focus on the joint optimization of near-field communication models and dynamic swarm formation in time-varying environments, thereby enabling promising functions such as integrated sensing and communication (ISAC)  in conjunction with DCB in LAE networks.




\vspace{-10 mm}

\begin{IEEEbiography}
[{\includegraphics[width=1in,height=1.25in,clip,keepaspectratio]{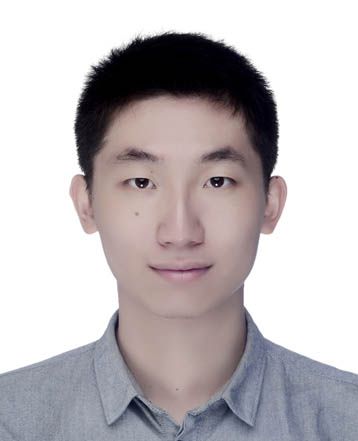}}]
{Jiahui Li} received his B.S. in Software Engineering, and M.S. and Ph.D. in Computer Science and Technology from Jilin University, Changchun, China, in 2018, 2021, and 2024, respectively. He was a visiting Ph.D. student at the Singapore University of Technology and Design (SUTD). He currently serves as an assistant researcher in the College of Computer Science and Technology at Jilin University. His current research focuses on integrated air-ground networks, UAV networks, wireless energy transfer, and optimization.
\end{IEEEbiography}

\vspace{-10 mm}

\begin{IEEEbiography}[{\includegraphics[width=1in,height=1.25in,clip,keepaspectratio]{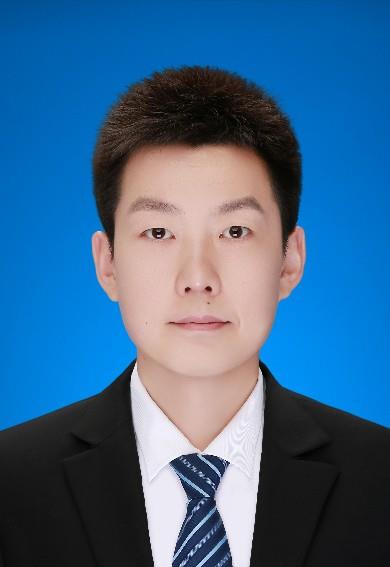}}]{Geng Sun} (Senior Member, IEEE) received the B.S. degree in communication engineering from Dalian Polytechnic University, and the Ph.D. degree in computer science and technology from Jilin University, in 2011 and 2018, respectively. He was a Visiting Researcher with the School of Electrical and Computer Engineering, Georgia Institute of Technology, USA. He is a Professor in College of Computer Science and Technology at Jilin University, and His research interests include wireless networks, UAV communications, collaborative beamforming and optimizations.
\end{IEEEbiography}

\vspace{-10 mm}

\begin{IEEEbiography}[{\includegraphics[width=1in,height=1.25in,clip,keepaspectratio]{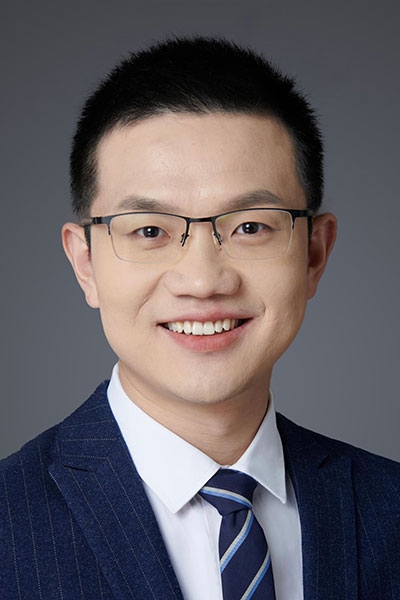}}]{Qingqing Wu}
(Senior Member, IEEE) is an Associate Professor with Shanghai Jiao Tong University. His current research interest includes intelligent reflecting surface (IRS), unmanned aerial vehicle (UAV) communications, and MIMO transceiver design. He has coauthored more than 100 IEEE journal papers with 30 ESI highly cited papers and 9 ESI hot papers, which have received more than 25,000 Google citations. He was listed as the Clarivate ESI Highly Cited Researcher since 2021, the Most Influential Scholar Award in AI-2000 by Aminer since 2021 and World’s Top 2\% Scientist by Stanford University in since 2020.

\par He was the recipient of the IEEE Communications Society Fred Ellersick Prize, IEEE  Best Tutorial Paper Award in 2023, Asia-Pacific Best Young Researcher Award and Outstanding Paper Award in 2022, Young Author Best Paper Award in 2021, the Outstanding Ph.D. Thesis Award of China Institute of Communications in 2017, the IEEE ICCC Best Paper Award in 2021, and IEEE WCSP Best Paper Award in 2015. He was the Exemplary Editor of IEEE Communications Letters in 2019 and the Exemplary Reviewer of several IEEE journals. He serves as an Associate Editor for \textsc{IEEE Transactions on Communications}, \textsc{IEEE Communications Letters}, \textsc{IEEE Wireless Communications Letters}. He is the Lead Guest Editor for \textsc{IEEE Journal on Selected Areas in Communications}. 
\end{IEEEbiography}

\begin{IEEEbiography}[{\includegraphics[width=1in,height=1.25in,clip,keepaspectratio]{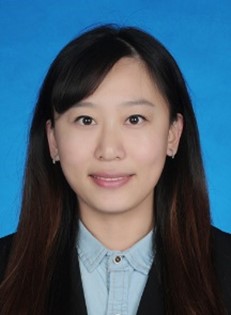}}]{Shuang Liang} received the B.S. degree in Communication Engineering from Dalian Polytechnic University, China in 2011, the M.S. degree in Software Engineering from Jilin University, China in 2017, and the Ph.D. degree in Computer Science from Jilin University, China, in 2022, respectively. She is currently postdoctoral researcher in School of Information Science and Technology, Northeast Normal University. Her research interests focus on wireless communication, design of array antennas, collaborative beamforming and optimizations.
\end{IEEEbiography}

\begin{IEEEbiography}[{\includegraphics[width=1in,height=1.25in,clip,keepaspectratio]{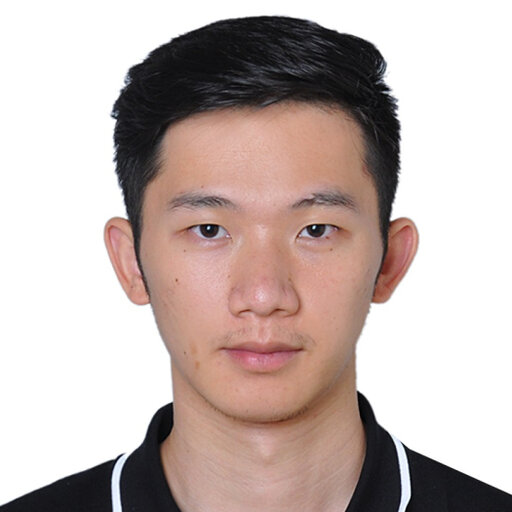}}]{Jiacheng Wang} received the Ph.D. degree from the School of Communication and Information Engineering, Chongqing University of Posts and Telecommunications, Chongqing, China. He is currently a Research Associate in computer science and engineering with Nanyang Technological University, Singapore. His research interests include wireless sensing, semantic communications, and metaverse.
\end{IEEEbiography}

\begin{IEEEbiography}[{\includegraphics[width=1in,height=1.25in,clip,keepaspectratio]{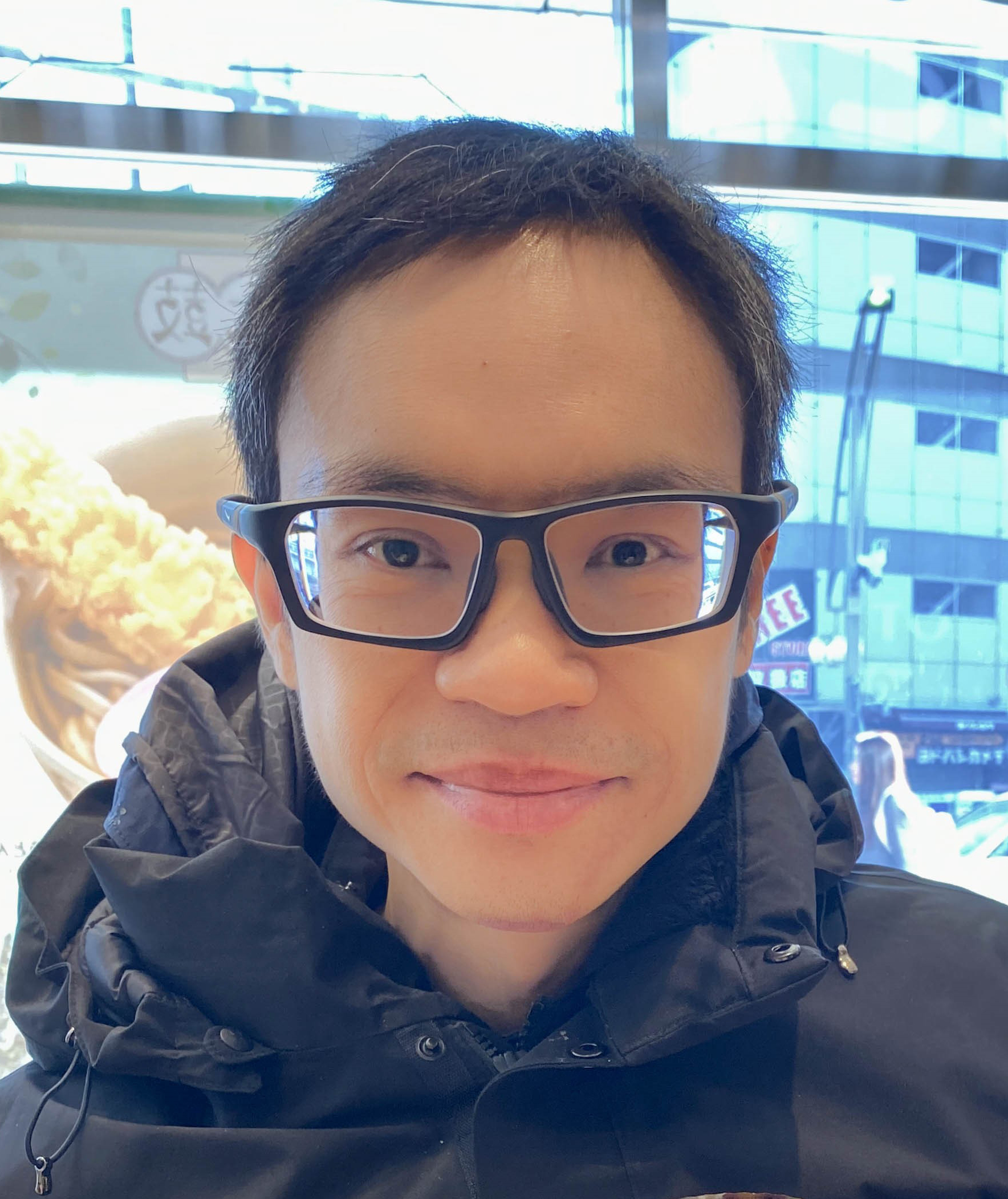}}]{Dusit Niyato} (Fellow, IEEE) is a professor in the College of Computing and Data Science, at Nanyang Technological University, Singapore. He received B.Eng. from King Mongkuts Institute of Technology Ladkrabang (KMITL), Thailand and Ph.D. in Electrical and Computer Engineering from the University of Manitoba, Canada. His research interests are in the areas of mobile generative AI, edge intelligence, decentralized machine learning, and incentive mechanism design.
\end{IEEEbiography}

\begin{IEEEbiography}[{\includegraphics[width=1in,height=1.25in,clip,keepaspectratio]{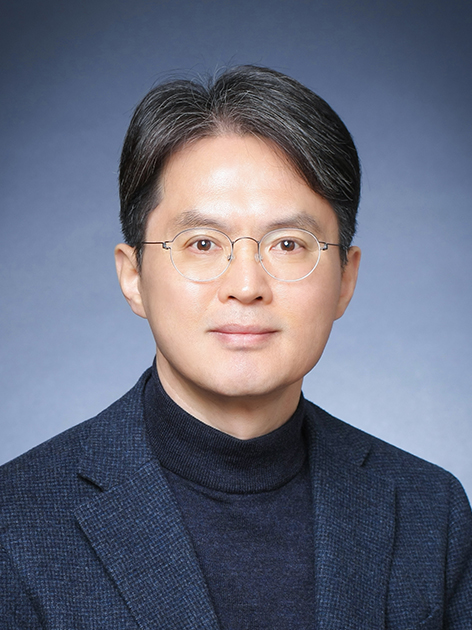}}]{Dong In Kim} (Fellow, IEEE) received the Ph.D. degree in electrical engineering from the University of Southern California, Los Angeles, CA, USA, in 1990. He was a Tenured Professor with the School of Engineering Science, Simon Fraser University, Burnaby, BC, Canada. He is currently a Distinguished Professor with the College of Information and Communication Engineering, Sungkyunkwan University, Suwon, South Korea. He is a Fellow of the Korean Academy of Science and Technology and a Member of the National Academy of Engineering of Korea. He was the first recipient of the NRF of Korea Engineering Research Center in Wireless Communications for RF Energy Harvesting from 2014 to 2021. He received several research awards, including the 2023 IEEE ComSoc Best Survey Paper Award and the 2022 IEEE Best Land Transportation Paper Award. He was selected the 2019 recipient of the IEEE ComSoc Joseph LoCicero Award for Exemplary Service to Publications. He was the General Chair of the IEEE ICC 2022, Seoul. Since 2001, he has been serving as an Editor, an Editor at Large, and an Area Editor of Wireless Communications I for \textsc{IEEE Transactions on Communications}. From 2002 to 2011, he served as an Editor and a Founding Area Editor of Cross-Layer Design and Optimization for \textsc{IEEE Transactions on Wireless Communications}. From 2008 to 2011, he served as the Co-Editor- in-Chief for the \textsc{IEEE/KICS Journal of Communications and Networks}. He served as the Founding Editor-in-Chief for the \textsc{IEEE Wireless Communications Letters} from 2012 to 2015. He has been listed as a 2020/2022 Highly Cited Researcher by Clarivate Analytics.
\end{IEEEbiography}

\end{document}